\documentclass[acmsmall]{acmart}
\usepackage[linesnumbered,ruled,vlined,algo2e]{algorithm2e}
\usepackage{makecell}
\usepackage{color}
\usepackage{multirow}
\usepackage{subcaption}
\usepackage[figuresright]{rotating}
\usepackage{wrapfig}
\usepackage{colortbl}
\newcommand{\etc}{{\em etc}}  
\definecolor{tblblue}{RGB}{129,184,223}
\definecolor{tblred}{RGB}{254,129,125}
\usepackage[normalem]{ulem}

\AtBeginDocument{%
  \providecommand\BibTeX{{%
    \normalfont B\kern-0.5em{\scshape i\kern-0.25em b}\kern-0.8em\TeX}}}

\setcopyright{acmlicensed}
\copyrightyear{2024}
\acmYear{2024}
\acmDOI{}

\acmSubmissionID{123-A56-BU3}
\begin{document}

\title{CLIP-DFGS: A Hard Sample Mining Method for CLIP in Generalizable Person Re-Identification}

\author{Huazhong Zhao}
\email{zhaohuazhong@seu.edu.cn}
\orcid{0009-0004-5667-0967}

\author{Lei Qi}
\authornotemark[1]
\email{qilei@seu.edu.cn}
\orcid{0000-0001-7091-0702}

\author{Xin Geng}
\email{xgeng@seu.edu.cn}
\orcid{0000-0001-7729-0622}
\affiliation{%
  \institution{School of Computer Science and Engineering, Southeast University, and Key Laboratory of New Generation Artificial Intelligence Technology and Its Interdisciplinary Applications (Southeast University), Ministry of Education}
  \city{Nanjing}
  \state{Jiangsu}
  \country{China}
  \postcode{210019}
}

\thanks{Corresponding author: Lei Qi}

\thanks{The work is supported by NSFC Program (Grants No. 62206052, 62125602, 62076063), China Postdoctoral Science Foundation (Grants No. 2024M750424), Supported by the Postdoctoral Fellowship Program of CPSF (Grant No. GZC20240252), and Jiangsu Funding Program for Excellent Postdoctoral Talent (Grant No. 2024ZB242).}

\renewcommand{\shortauthors}{Zhao and Qi, et al.}

\begin{abstract}
Recent advancements in pre-trained vision-language models like CLIP have shown promise in person re-identification (ReID) applications.
However, their performance in generalizable person re-identification tasks remains suboptimal.
The large-scale and diverse image-text pairs used in CLIP's pre-training may lead to a lack or insufficiency of certain fine-grained features.
In light of these challenges, we propose a hard sample mining method called DFGS (Depth-First Graph Sampler), based on depth-first search, designed to offer sufficiently challenging samples to enhance CLIP's ability to extract fine-grained features.
DFGS can be applied to both the image encoder and the text encoder in CLIP.
By leveraging the powerful cross-modal learning capabilities of CLIP, we aim to apply our DFGS method to extract challenging samples and form mini-batches with high discriminative difficulty, providing the image model with more efficient and challenging samples that are difficult to distinguish, thereby enhancing the model's ability to differentiate between individuals.
Our results demonstrate significant improvements over other methods, confirming the effectiveness of DFGS in providing challenging samples that enhance CLIP's performance in generalizable person re-identification.
\end{abstract}




 \received{31 July 2024}
 \received[revised]{28 September 2024}
 \received[accepted]{13 October 2024}

\maketitle

\section{Introduction}
In response to the growing demand for accurate person matching across unseen domains, generalizable person re-identification (DG-ReID) has emerged as a particularly prominent research topic~\cite{Liao_Shao_2020, Liao_Shao, He_Liu_Liang_Zheng_Liao_Cheng_Mei, Zhang_Dou_Yunlong_Li, xu2022mimic, Liao_Shao_2021}.
This area of research has a large and promising practical application space, addressing critical needs in various real-world scenarios~\cite{Liao_Shao,2023Dual,yan2021beyond,yang2020part,Liu2022GenerativeML,10.1145/3419439}.
It has gained significant attention in recent years due to its widespread and essential applications in public security, and human tracking systems~\cite{Zheng_Yang_Hauptmann_2016, Ye_Shen_Lin_Xiang_Shao_Hoi_2022, Leng_Ye_Tian_2020,2020PurifyNet}.

\begin{figure}[ht]
    \begin{subfigure}{0.32\linewidth}
    \centering
    \includegraphics[width=0.99\linewidth]{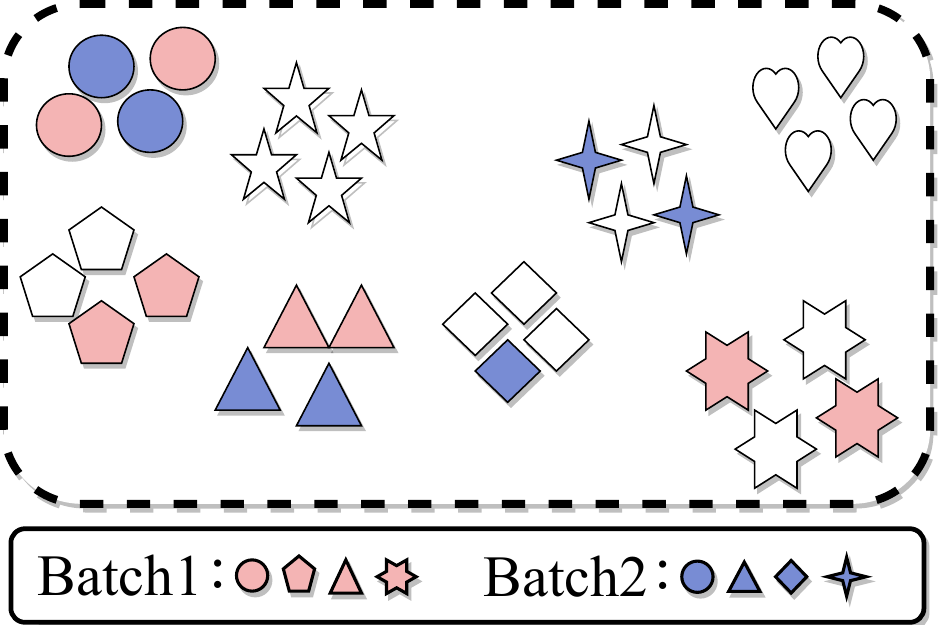}
    \caption{}
    \label{fig:1a}
    \end{subfigure}
\hfill
  \begin{subfigure}{0.32\linewidth}
    \centering
    \includegraphics[width=0.99\linewidth]{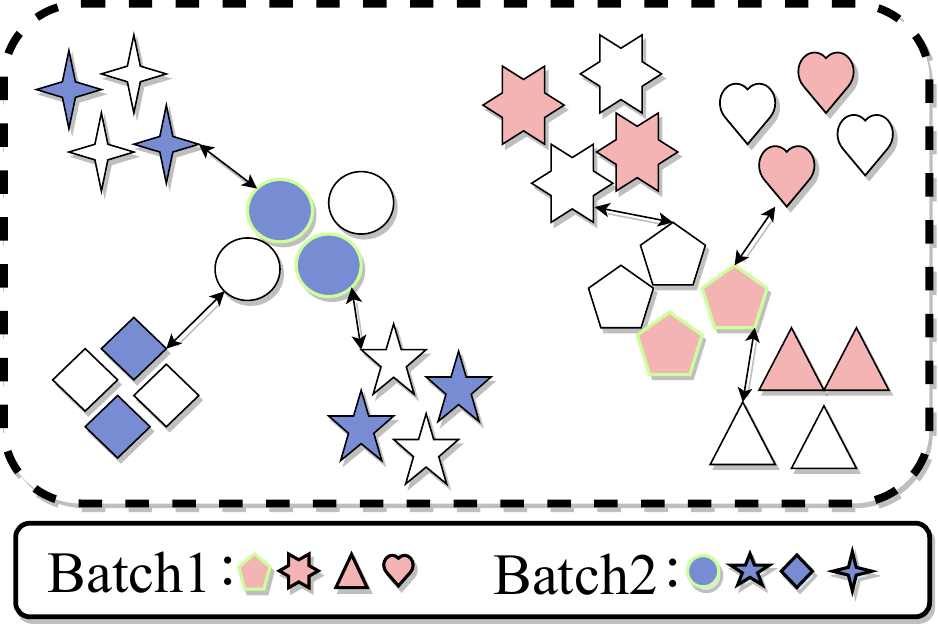}
    \caption{}
    \label{fig:1b}
\end{subfigure}
\hfill
  \begin{subfigure}{0.32\linewidth}
    \centering
    \includegraphics[width=0.99\linewidth]{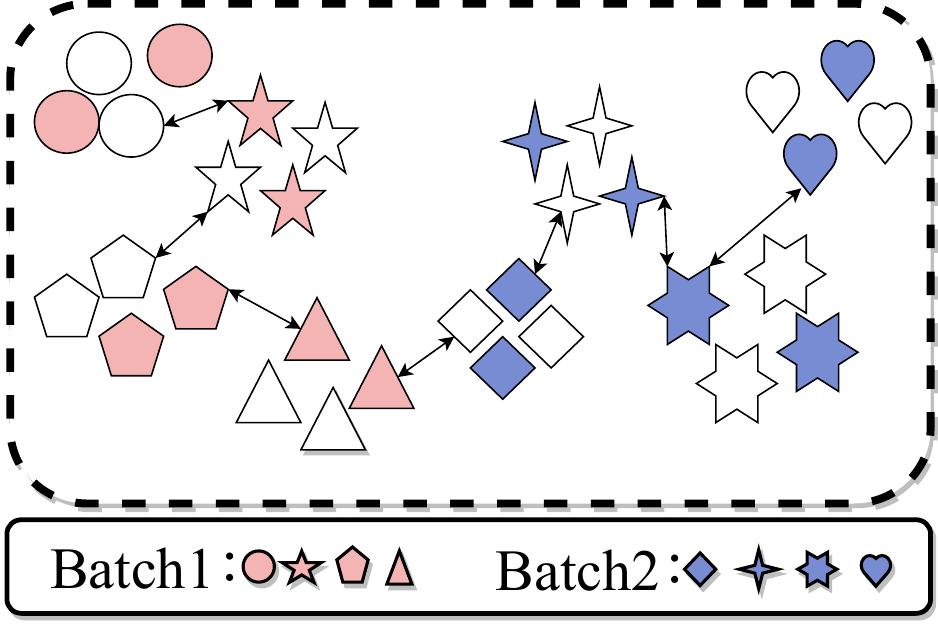}
    \caption{}
    \label{fig:1c}
\end{subfigure}
    \caption{Three different sampling methods: (a) PK sampler; (b) Graph Sampler (GS); (c) Depth First Graph Sampler (DFGS). Each shape represents a different class, and each color represents a different batch.}
 \label{fig:1}

\end{figure}

Recent advancements in pre-trained vision-language models, such as CLIP (Contrastive Language-Image Pre-Training)~\cite{radford2021learning}, have shown great promise in enhancing ReID applications~\cite{chen2018improving,farooq2020convolutional,li2023clip}.
CLIP's ability to understand both visual and textual data through cross-modal learning makes it a powerful tool for this purpose.
This unique capability positions CLIP as a strong candidate for improving ReID performance.
Despite its potential, CLIP's performance in generalizable ReID tasks is suboptimal.
A significant contributing factor is the insufficient representation of fine-grained features, which hinders the model's ability to effectively differentiate between challenging instances~\cite{yan2023clip,radford2021learning}.
The original CLIP model, pretrained on extensive datasets, exhibits lower performance on fine-grained tasks compared to ResNet50 and struggles with more complex challenges.
Person Re-Identification, which requires fine-grained feature extraction, exemplifies such difficulties~\cite{yan2023clip}.

Traditional sampling methods, such as the commonly used PK sampler~\cite{Hermans_Beyer_Leibe_2017}, involve randomly selecting \(P\) classes, with each class randomly choosing \(K\) samples to form a mini-batch shown in Fig.~\ref{fig:1a}.
While this method provides a diverse set of samples, it fails to consistently offer sufficiently challenging samples that can enhance the model's learning capabilities~\cite{Liao_Shao_2021}.
This limitation hinders the model's ability to generalize well to unseen instances, which is crucial for generalizable person re-identification performance.
The PK sampler's inability to focus on hard samples results in a training process that may not fully exploit the potential of the model.
Although Liao et al.\cite{Liao_Shao_2021} proposed the GS method, which mines hard samples for each class from the entire training set for generalizable person re-identification (as shown in Fig.\ref{fig:1b}), GS is specifically designed for use with their metric networks~\cite{Liao_Shao_2020,Liao_Shao}.

\begin{figure*}[t]
\centering
\includegraphics[width=0.88\columnwidth]{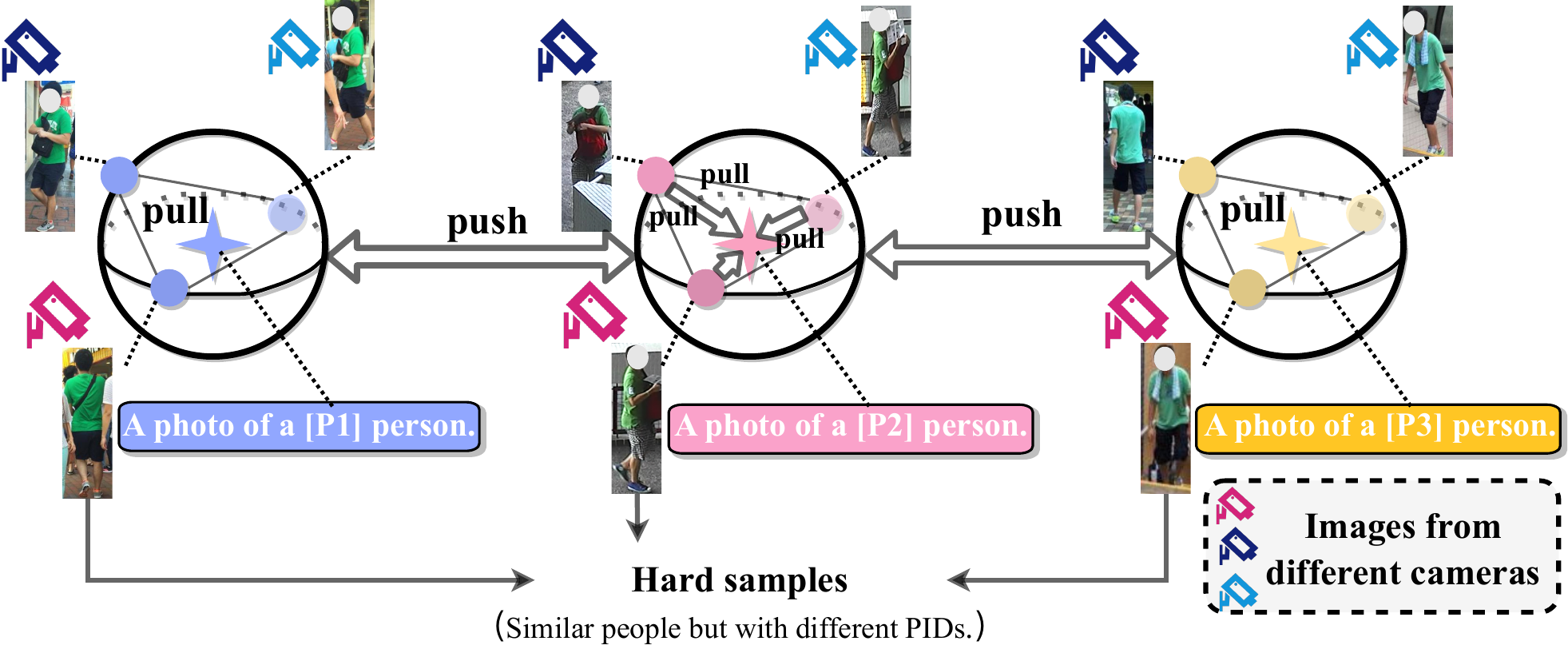}
\caption{On one hand, the same individual captured by different cameras may exhibit significant differences due to variations in angles, backgrounds, resolutions, \etc. Thus, we define individuals captured by different cameras as intra-class hard samples. On the other hand, in the training dataset, there may exist samples that are very similar but do not belong to the same individual, so we define these as inter-class hard samples.}
\label{fig:2}
\end{figure*}

In light of these challenges, we propose an efficient mini-batch sampling method called the Depth First Graph Sampler (DFGS) to enhance CLIP's ability to extract fine-grained features.
This method utilizes a depth-first search algorithm on the constructed graph to form mini-batches composed of hard samples, as shown in Fig.~\ref{fig:1c}, more details are provided in Fig.~\ref{fig:2}
By doing so, it can provide informative and challenging samples that enhance the model's learning process.
Although the GS method selects neighboring nodes for each node to create a batch, it cannot inherently guarantee that the batch will be densely populated with challenging samples.
Through DFGS, we can more effectively supply valuable samples for the model's training.
Moreover, by utilizing the cross-modal learning capabilities of CLIP, our DFGS method extracts challenging samples and forms mini-batches with high discriminative difficulty.
Furthermore, DFGS can be applied to both the image encoder and the text encoder in CLIP, enabling a more comprehensive approach to hard sample mining.
By focusing on hard samples, our method ensures that the model is trained on the most challenging samples, leading to better generalization and improved performance in DG-ReID tasks.
This method ensures that the image model is trained with samples that are difficult to distinguish, thereby enhancing its ability to differentiate between individuals.

In this paper, we delve into the details of the DFGS method, its implementation within the CLIP framework, and the empirical validation of its efficacy.
We aim to provide a comprehensive understanding of the DFGS method and its implementation within the CLIP framework.
Detailed methods and comprehensive experimental analyses are provided in the following sections.
In summary, our main contributions are summarized as follows:

\begin{enumerate}
\item[$\bullet$] We propose a novel sampling method called Depth First Graph Sampler (DFGS) and convincingly prove its remarkable efficacy in metric learning.
\item[$\bullet$] Based on the characteristics of CLIP, we propose specific DFGS sampling methods for the image encoder and text encoder respectively.
\item[$\bullet$] Extensive experiments on several standard benchmark datasets show that our method can achieve significant improvements in generalizable person re-identification.
\end{enumerate}

\section{Related Work}
In this section, we conduct an extensive investigation into some of the most relevant works, aiming to offer a detailed overview and present a summary of the most relevant works.

\subsection{Generalizable Person Re-identification}
The goal of generalizable person re-identification is to learn a model in the source domain that can directly perform well in the target domain without additional training.
Generalizable person re-identification necessitates learning features with high discriminative power to accurately identify individuals across different environments.
However, conventional domain generalization methods may not be sufficient for developing feature representations specifically tailored to person re-identification.
Existing methods mainly include network normalization~\cite{DBLP:conf/nips/EomH19,DBLP:conf/bmvc/JiaRH19,kale2023provenance,DBLP:conf/cvpr/JinLZ0Z20,luo2019strong,xu2022mimic}, meta-learning~\cite{DBLP:conf/icpr/LinCW20,DBLP:conf/cvpr/SongYSXH19,DBLP:conf/cvpr/ZhaoZYLLLS21,DBLP:conf/cvpr/DaiLLTD21}, and domain alignment~\cite{DBLP:conf/aaai/ChenDLZX0J21,DBLP:conf/eccv/LuoSZ20,DBLP:conf/wacv/YuanCCYRW020,DBLP:conf/eccv/ZhuangWXZZWAT20,DBLP:conf/eccv/LiaoS20,DBLP:conf/iccv/ZhuPIE17,2020Attribute,qi2024multimatch,fang2023three}.
This is because person re-identification presents unique challenges, such as variations in lighting, pose, and occlusions, which require specialized solutions~\cite{2019Person}.

As a result, researchers often need to create customized methods and techniques that address the specific characteristics of person re-identification tasks.
These tailored approaches enhance the model's ability to accurately re-identify individuals across diverse surveillance scenarios, improving performance in real-world applications.

For instance, the META framework~\cite{xu2022mimic} considers the relevance of target samples and source domains through normalization statistics.
It incorporates an aggregation module to dynamically combine multiple experts~\cite{dai2021generalizable}, enabling the model to adapt to the characteristics of the unseen target domain effectively.
Similarly, the ACL framework~\cite{Zhang_Dou_Yunlong_Li} is augmented with a Cross-Domain Embedding Block (CODE-Block).
This component ensures a shared feature space that captures both domain-invariant and domain-specific features. The CODE-Block also dynamically explores relationships across different domains, facilitating a more robust learning process.

These innovative methods demonstrate the importance of developing specialized techniques in generalizable person re-identification, highlighting the need for continuous research and development to tackle the evolving challenges in this field.

\subsection{Vision-language Models in Person Re-Identification}
Vision-language models in person re-identification have shown significant promise, with CLIP-ReID~\cite{li2023clip} being a notable contribution leveraging the CLIP framework.
CLIP-ReID employs a two-stage strategy to enhance visual representation.
The central concept involves maximizing the cross-modal descriptive capability inherent in CLIP by utilizing a set of trainable text tokens for each individual identity.
This innovative method allows the model to effectively bridge the gap between visual and textual modalities, resulting in more accurate and generalizable person re-identification.

The application of Vision-Language Models (VLMs) in Text-to-Image Person Re-Identification~\cite{yan2023clip,jiang2023cross} is particularly extensive.
This approach focuses on associating textual descriptions with corresponding images for individual recognition. Text-to-Image Person ReID benefits greatly from advanced pre-training techniques, drawing inspiration from the successful application of vision-language models. These models, trained on extensive cross-modal datasets, have demonstrated remarkable proficiency in learning intricate associations between images and text, enabling them to perform well even in complex real-world scenarios where descriptions and visual appearances can vary significantly.

Researchers are increasingly exploring the adaptation of such models to the specific task of ReID with textual descriptions~\cite{li2023clip,jiang2023cross,Li_Xiao_Li_Zhou_Yue_Wang_2017,Zhu_Wang_Li_Wan_Jin_Wang_Hu_Hua_2021,ding2021semantically,Wang_Zhu_Xue_Wan_Liu_Wang_Li_2022}.
These efforts include developing methods to fine-tune pre-trained vision-language models on person re-identification datasets, improving the alignment between textual and visual features.
Moreover, there is ongoing work to enhance the robustness of these models against variations in lighting, pose, and occlusions, which are common challenges in person re-identification tasks.

By integrating textual descriptions, vision-language models can leverage additional contextual information that purely visual models might miss.
This multimodal approach~\cite{zheng2022faster} not only improves identification accuracy but also provides a more comprehensive understanding of the data.
As a result, vision-language models represent a promising direction for future research and development in person re-identification, with the potential to significantly advance the field and improve the performance of surveillance and security systems.

\subsection{Hard Sample Mining in Person Re-Identification}
In person re-identification, samplers play a crucial role in training effective models~\cite{Hermans_Beyer_Leibe_2017,zhang2021one}.
The most commonly used method is the PK sampler~\cite{Hermans_Beyer_Leibe_2017}.
This method randomly selects \( P \) classes and then samples \( K \) images from each class, forming a mini-batch of size \( B = P  \times  K \).
While the PK sampler is straightforward and widely adopted, its entirely random execution might not always provide the most informative and challenging samples needed for effective metric learning in person re-identification.

To address this limitation, more sophisticated sampling methods have been explored.
One notable recent method is the Graph Sampler (GS)~\cite{Liao_Shao_2021}.
This method has been utilized in models such as QAconv$_{50}$~\cite{Liao_Shao_2020} and TransMatcher~\cite{Liao_Shao}.
The GS employs binary cross-entropy loss to measure pair-wise sample distances, aiming to construct a nearest neighbor relationship graph for all classes at the beginning of each training epoch.
The main objective of the GS is to strategically sample by ensuring that each mini-batch comprises a randomly chosen class along with its top-k nearest neighboring classes.
This method effectively introduces more challenging and informative samples into the training process.
By focusing on the nearest neighbors, the GS can provide the model with samples that are more likely to be difficult, thereby enhancing the discriminative power of the learned feature representations.

The GS method demonstrates that strategic sampling can significantly impact the training efficiency and performance of person re-identification models.
By carefully selecting samples that present a higher degree of difficulty, these advanced sampling methods can facilitate better metric learning, ultimately leading to more accurate and reliable person re-identification systems.
This highlights the ongoing need for innovative sampling strategies in the development of robust re-identification models.

\section{Method}
\label{sec:method}
\begin{figure*}[t]
  \centering
    \includegraphics[width=0.98\columnwidth]{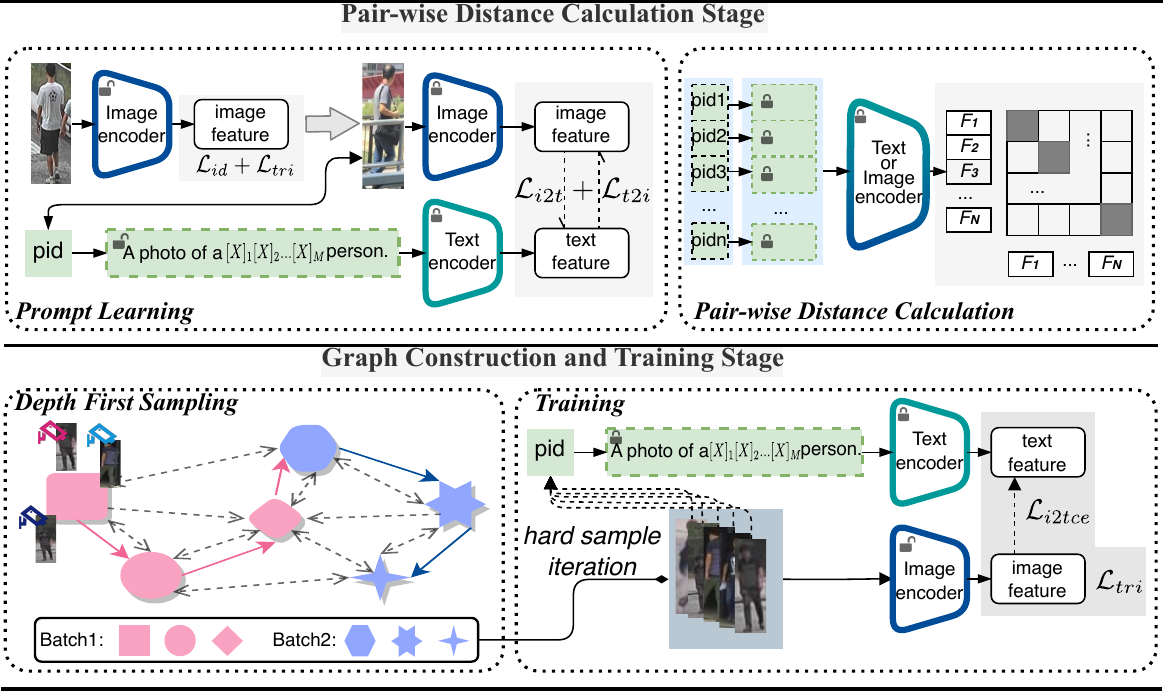}
   \caption{Overview of our method. Firstly, for each person ID, a specific text description is learned. Then, based on the acquired text descriptions, features are extracted and the pairwise distance similarity matrix is calculated and saved. Subsequently, during the sampling and learning stages, a sample graph is constructed using the pairwise distance similarity matrix. Through a depth-first search on this sample graph, training sample iterations are obtained, thus providing mini-batches containing challenging samples for fine-tuning the image encoder. Here, we use a directed graph for representing the structure, dashed lines indicate the directed edges of the graph, while solid lines represent the traversal sequence of the depth-first search.}
  \label{fig:3}
\end{figure*}

In this section, we introduce the details of the proposed method.
As shown in Fig.~\ref{fig:3}, we first employ an image encoder to learn text prompts, following the CLIP-ReID~\cite{li2023clip}.
We then retain the pair-wise distance matrix between the features of text prompts with different pids for use in subsequent sampling.
Subsequently, we propose the Depth First Graph Sampler (DFGS) during the sampling stage, facilitating the inclusion of challenging samples within a batch.

\subsection{Preliminaries}
\label{sec:preliminaries}

\textbf{CLIP.}
Contrastive Language-Image Pre-training (CLIP)~\cite{radford2021learning} is a model developed by OpenAI that learns visual concepts from natural language descriptions.
CLIP consists of an image encoder (e.g., a Vision Transformer or ResNet) and a text encoder (Transformer-based) that map images and text into a shared embedding space.
The model is trained using a contrastive learning objective, which maximizes the cosine similarity between matching image-text pairs and minimizes it for non-matching pairs. The training objective is defined as:

\begin{align}
    \mathcal{L} = & -\frac{1}{2N} \sum_{i=1}^{N} \Bigg[ \log \frac{\exp(\text{s}(I_i, T_i)/\tau)}{\sum_{j=1}^{N} \exp(\text{s}(I_i, T_j)/\tau)} + \log \frac{\exp(\text{s}(T_i, I_i)/\tau)}{\sum_{j=1}^{N} \exp(\text{s}(T_i, I_j)/\tau)} \Bigg],
\end{align}
where $\text{s}(x, y)$ represents the cosine similarity between $x$ and $y$, and $\tau$ is a temperature parameter.

\textbf{CLIP-ReID.}
To address the limitations in textual information for person or vehicle re-identification, CLIP-ReID is proposed.
This method builds on the pre-trained CLIP model and involves two training stages, significantly improving performance compared to the baseline.

In the first training stage, ID-specific learnable tokens are introduced to capture ambiguous text descriptions for each ID independently. The text descriptions are structured as ``A photo of a \([X]_1\) \([X]_2\) \([X]_3\) \ldots \([X]_M\) person/vehicle'', where each \( [X]_M \) is a learnable token of the same dimension as word embeddings, and M is the number of tokens.
During this stage, the parameters of the image encoder \( I(\cdot) \) and text encoder \( T(\cdot) \) are fixed, and only the tokens [X]m are optimized. The loss functions \( \mathcal{L}_{i2t} \) and \( \mathcal{L}_{t2i} \) are adapted to replace \( \text{text}_i \) with \( \text{text}_{y_i} \), as each ID shares the same description. Specifically, \( \mathcal{L}_{t2i} \) is modified to:
\begin{align}
    \mathcal{L}_{t2i}(y_i) = -\frac{1}{|P(y_i)|} \sum_{p \in P(y_i)} \log \frac{\exp(s(V_p, T_{y_i}))}{\sum_{a=1}^{B} \exp(s(V_a, T_{y_i}))},
\end{align}
where \( P(y_i) \) is the set of indices of all positives for \( T_{y_i} \) in the batch, and \( |\cdot| \) is its cardinality. \( s(x, y) \) denotes the similarity score between two vectors \( x \) and \( y \), typically the cosine similarity. By minimizing \( \mathcal{L}_{i2t} \) and \( \mathcal{L}_{t2i} \), gradients are back-propagated through the fixed \( T(\cdot) \) to optimize the tokens, fully leveraging \( T(\cdot) \). The total loss for this stage is:
\begin{align}
    \mathcal{L}_{\text{stage1}} = \mathcal{L}_{i2t} + \mathcal{L}_{t2i}.
\end{align}
To improve efficiency, all image features are extracted at the beginning of this stage, and different \( T_{y_i} \) for all IDs are saved for the next stage.

In the second training stage, only the parameters of \( I(\cdot) \) are optimized. A strong ReID pipeline is followed using triplet loss \( \mathcal{L}_{\text{tri}} \) and ID loss \( \mathcal{L}_{\text{id}} \) with label smoothing, defined as:
\begin{align}
    \mathcal{L}_{\text{id}} &= \sum_{k=1}^{N} -q_k \log(p_k), \\
    \mathcal{L}_{\text{tri}} &= \max(d_p - d_n + \alpha, 0),
\end{align}
where \( q_k \) is the smoothed label for class \( k \), \( p_k \) is the predicted probability for class \( k \), \( d_p \) and \( d_n \) are the distances between features of positive and negative pairs, respectively, and \( \alpha \) is the margin for the triplet loss. Additionally, text features obtained in the first stage are used to calculate the image-to-text cross-entropy \( \mathcal{L}_{i2tce} \) with label smoothing:
\begin{align}
    \mathcal{L}_{i2tce}(i) = \sum_{k=1}^{N} -q_k \log \frac{\exp(s(V_i, T_{y_k}))}{\sum_{y_a=1}^{N} \exp(s(V_i, T_{y_a}))}.
\end{align}
The total loss for the second stage is:
\begin{align}
    \mathcal{L}_{\text{stage2}} = \mathcal{L}_{\text{id}} + \mathcal{L}_{\text{tri}} + \mathcal{L}_{i2tce}.
\end{align}
The entire training process leverages learnable prompts to effectively capture and store the hidden states of the pre-trained encoders, thereby preserving the inherent advantages of CLIP.
These learnable prompts serve as a bridge, maintaining the rich information embedded in the pre-trained models. During the second stage of training, these prompts play a crucial role in regularizing the image encoder.
This regularization process not only stabilizes the training but also significantly enhances the generalization capabilities of the model, ensuring it performs well across diverse and unseen data.

\subsection{Pair-wise Distance Calculation Stage}
\textbf{Prompt Learning.}
Firstly, we refer ot CLIP-ReID~\cite{li2023clip} to introduce learnable tokens specific to IDs to learn textual descriptions that are independent of each ID.
Before commencing the learning of prompts, we update \( I(\cdot) \) using ID loss and Triplet loss to ensure the prompts learned are more accurate:
\begin{eqnarray}
    \mathcal{L}_{pre} = \mathcal{L}_{id} + \mathcal{L}_{tri}.
\end{eqnarray}

The textual description input to \( T(\cdot) \) is designed as ``a photo of a\([X]_1\) \([X]_2\) \([X]_3\) \ldots \([X]_M\) person''.
\( M \) represents the number of learnable text tokens, which we set to 4 following the configuration of CLIP-ReID.
Similar to CLIP-ReID, we fix the parameters of \( I(\cdot) \) and \( T(\cdot) \), and only optimize the tokens \([X]_m\) (\( m \in 1, \ldots, M \)) and utilize \( \mathcal{L}_{i2t} \) and \( \mathcal{L}_{t2i} \), the training loss for prompt learning is represented as:
\begin{eqnarray}
    \mathcal{L}_{PL} = \mathcal{L}_{i2t} + \mathcal{L}_{t2i}.
\end{eqnarray}

%\begin{eqnarray}
%\mathcal{L}_{t2i}(y_i) = -\frac{1}{|P(y_i)|}\sum_{p \in P(y_i)} \log \frac{\exp(s(V_p, T_{y_i}))}{\sum_{a=1}^{B} \exp(s(V_a, T_{y_i}))},
%\end{eqnarray}

\textbf{Pair-wise Distance Calculation.}
Before proceeding with sampling and training, we extract features from the learned text encoder or image encoder and compute the euclidean distance between each pair:
\begin{eqnarray}
D_{\text{pair}}(F)  = \begin{pmatrix}
\infty & s(F_1,F_2) &s(F_1,F_3) &  \ldots & s(F_1,F_N) \\
s(F_2,F_1) & \infty &s(F_2,F_3) &  \ldots & s(F_2,F_N) \\
s(F_3,F_1) & s(F_3,F_2) &\infty &  \ldots & s(F_3,F_N) \\
\ldots &  \ldots & \ldots & \ldots & \ldots \\
s(F_N,F_1) & s(F_N,F_2) & s(F_N,F_3) & \ldots & \infty
\end{pmatrix},
    \label{dist}
\end{eqnarray}
where \(F_n (n \in 1, 2, \ldots, N)\) represents the features extracted from Text encoder \( T(\cdot)\) or Image encoder \( I(\cdot)\).
Here, \(N\) denotes the number of IDs (the number of persons).
\( s(\cdot) \) denotes the distance calculation method, where in this paper, we utilize the Euclidean distance.
Moreover, the rationale behind setting the diagonal to \(\infty\) is to prevent the sampling stage from gathering samples of oneself. This is because we classify instances with smaller distance values as hard samples, and hard samples must belong to different IDs.

\subsection{Graph Construction and Training Stage}

\textbf{Graph Construction and Depth First Sampling.}
The well-known PK sampler~\cite{Hermans_Beyer_Leibe_2017}, due to its fully random execution, is likely to be insufficient in providing informative and efficient samples for person re-identification metric learning~\cite{Liao_Shao_2021}.
In response to this problem, GS~\cite{Liao_Shao_2021} selects neighboring nodes for each node to create a batch, as shown in Fig.~\ref{fig:3}.
However, this method cannot inherently guarantee that the batch will be densely populated with challenging samples.
So we propose an efficient mini-batch sampling method, named Depth First Graph Sampler (DFGS).
First and foremost, it's crucial to note that our method differs from samplers like GS~\cite{Liao_Shao_2021}, which solely consider image features for hard sample mining.
In order to enhance the model's discriminative capabilities for fine-grained features, distinguishing similar yet distinct individuals, we consider composing mini-batches with different samples that have similar features.

\begin{algorithm2e}[!ht]

\DontPrintSemicolon

\caption{\textbf{Depth-First Graph Sampler}}
\label{alg:1}
\SetKwInOut{Input}{input}\SetKwInOut{Output}{output}
\Input{Smaple graph \(G\), batch size \(B\), number of instances per class \(n\) }
\Output{Sample iterator}
final\_idxs = []

batch\_idxs = []

stack = [random.choice(pids, size=1)]

\For {p in pids}{
shuffle(\(G\)[p])\textcolor{blue} {\tcp{Randomize node order to increase iteration diversity.}}
}

\While {stack is not empty }{
p = stack.pop()\textcolor{blue} {\tcp{depth-first search}}
\If{p is not available}{
continue
}

batch\_idxs.extend(diff\_cid\_sample(p,\(n\)))\textcolor{blue} {\tcp{Same person from different cameras.}}

\If{len (batch\_idxs) == \(B\)}{
        final\_idxs.extend(batch\_idxs)

        batch\_idxs = []
    }
\For {v in \(G\)[p][::-1]}{
    \If{v is available}{
        stack.append(v)
        }
    }
}
\Return iter(final\_idxs)

\end{algorithm2e}

Similar to GS, we construct a graph for all classes with both out-degree and in-degree set to \( K\), facilitating efficient depth-first sampling, as shown in Alg.~\ref{alg:1}.
Subsequently, we construct a graph for all classes with both out-degree and in-degree set to \(K\), facilitating efficient depth-first sampling, as shown in Alg.~\ref{alg:1}.
Next, for each class \( p\), we can use the pair-wise distance metirc \(D_{\text{pair}}\)~\eqref{dist} to obtain its top-K nearest classes, denoted as:
\begin{eqnarray}
G[p]= \{x_{p}^{i} | i = 1, 2, \ldots, K\},
    \label{gp}
\end{eqnarray}
where \( K\) represents the in-degree and out-degree of the graph.
Leveraging this information, we could effectively construct a graph \( G = (V, E)\) structure, wherein \( V = \{p|p= 1, 2, \ldots, N\}\) denotes the vertices and \( E = \{ (p_1, p_2)|p_2 \in G[p_1]\}\) represents the edges within the graph structure.

Furthermore, to control the difficulty level of hard samples, we do not simply select the top-k as the hard samples for the current sample.
Instead, we introduce a difficulty coefficient \( m \) and the number of hard samples \( k \).
Therefore, the equation for Eq.~\eqref{gp} can be modified to:
\begin{eqnarray}
G[p]= \{x_{p}^{i} | i = m+1, m+2, \ldots, m+k\},
    \label{gpv2}
\end{eqnarray}
where \( m \) and \( k \) serve as hyperparameters that effectively control the difficulty level and the number of samples in the sampling process.
Further investigation on hyperparameters is discussed in Sec.~\ref{sec:param}.

Furthermore, it is noted in the Alg.~\ref{alg:1} that we perform a shuffle operation on the graph nodes, denoted as ``shuffle(G[p])''.
This is because we believe that as training becomes more proficient, if some randomness is not introduced, each iteration within each epoch may become highly similar.
Specifically, the same challenging sample pairs would frequently end up in the same mini-batch, which obviously hampers the model's generalization ability.
Therefore, we shuffle the nodes of the graph, as indicated in Eq.~\eqref{gpv2}.
This way, the stack used in the subsequent depth-first sampling process exhibits significant randomness, thereby enhancing the diversity of iterations.
This research is addressed in Sec.~\ref{sec:abs}.

The final sampling method, for a randomly chosen class \( p\), we conduct a depth-first search on sample graph \( G\) employing a stack.
It should be noted that within a batch, the number of samples for a class \( p\) can only be \( n\).
Therefore, we check if there are already \( n\) samples for class \( p\) in the current batch.
If so, we continue the while loop by popping the next \( p\) from the stack.
Finally, for the hard samples within the class, we select \( n\) samples that are from as many different cameras as possible under class \( p\).
This not only leads to the incorporation of \textbf{inter-class} challenging samples within a mini-batch but also ensures that the \( n\) samples comprising a class are \textbf{intra-class} challenging, given the substantial dissimilarity in perspectives and styles across different camera views.
Thus, we can obtain an iteration based on the depth-first search.

%What's more, it is important to note that, due to the pre-computation and storage of this distance matrix, the subsequent sampling and training significantly improve time efficiency, thereby optimizing overall performance.

\textbf{Fine Tuning of \( I(\cdot) \).}
Through the aforementioned process, we can obtain mini-batches of challenging samples via depth-first search on the constructed graph.
We employ triplet loss to constrain metric learning, reducing the distance between samples of the same identity while pushing apart samples from different identities.
Additionally, we incorporate the \(\mathcal{L}_{i2tce}\) function from CLIP-ReID to jointly compose the loss function for fine-tuning the image encoder:
\begin{eqnarray}
\mathcal{L}_{\text{dfgs}} = \mathcal{L}_{tri} + \mathcal{L}_{i2tce} .
\label{loss}
\end{eqnarray}

\textbf{Remark 1:}
PK~\cite{Hermans_Beyer_Leibe_2017} randomly selects classes to form a batch, without considering their relationship with other classes in the overall training set.
GS~\cite{Liao_Shao_2021} simply selects the top-k nearest neighboring classes from the overall training set for each class to form an independent batch, resulting in the selected top-k nearest neighboring classes being similar.
In contrast, our method utilizes the depth-first search algorithm to fill a batch with as many neighboring challenging triplets as possible, thereby optimizing overall performance.
Moreover, we find that incorporating ID loss with triplet loss for extracted features does not lead to significant improvements in our task.
Therefore, we solely apply triplet loss for extracted features during the stage of fine-tuning the image encoder because our designed DFGS method is specifically tailored for triplet loss.

\textbf{Remark 2:}
Although DFGS can be applied to both the image encoder and the text encoder, each approach has advantages.
When applied to the image encoder, the DFGS\(_{I(\cdot)}\) sampler can fully exploit hard samples within the current epoch, making each round of training as targeted as possible.
When DFGS is applied to the text encoder (DFGS\(_{T(\cdot)}\)), it leverages text features to provide a comprehensive semantic understanding, thereby avoiding the dilemma of choosing which image to use as the class representative.
Furthermore, due to the pre-computation and storage of the pair-wise distance matrix, subsequent sampling and training significantly improve time efficiency.

\section{Experiments}
\label{sec:experiments}

\subsection{Experimental Settings}
\textbf{Datasets.}
We conduct extensive experiments on nine widely recognized public person re-identification (ReID) datasets, namely Market1501~\cite{zheng2015scalable}, MSMT17~\cite{Zheng_Shen_Tian_Wang_Wang_Tian_2015}, CUHK02~\cite{Li_Wang_2013}, CUHK03~\cite{Li_Zhao_Xiao_Wang_2014}, CUHK-SYSU~\cite{Xiao_Li_Wang_Lin_Wang_2016}, PRID~\cite{Hirzer_Beleznai_Roth_Bischof_2011}, GRID~\cite{Loy_Xiang_Gong_2010}, VIPeR~\cite{Gray_Tao_2008}, and iLIDs~\cite{Zheng_Gong_Xiang_2009}. These datasets vary in terms of the number of images, number of identities, and the complexity of capturing conditions, providing a comprehensive testbed for evaluating ReID models.
We evaluate our method using the Cumulative Matching Characteristics (CMC) and mean average precision (mAP) metrics, which are the standard evaluation protocols for person re-identification.
These metrics provide a detailed understanding of both the ranking performance and the precision of the model across different datasets.
To simplify our discussions, we use abbreviations to denote the datasets: Market1501 as M, MSMT17 as MS, CUHK02 as C2, CUHK03 as C3, and CUHK-SYSU as CS.

\textbf{Experimental protocols.}
\label{sec:protocols}
We follow three distinct protocols to assess the generalization capability of our model across multiple domains.
Under Protocol-1, the model is trained on a combination of the Market1501, CUHK02, CUHK03, and CUHK-SYSU datasets (M+C2+C3+CS).
The trained model is then tested on four separate datasets (PRID, GRID, VIPeR, and iLIDs) to evaluate its generalization to unseen domains.
Protocol-2 involves a single-domain testing approach where one dataset (M, MS, CS, or C3) is reserved for testing, while the remaining datasets are used for training.
This approach helps to understand how well the model trained on multiple sources performs when tested on a single unseen domain.
Protocol-3 closely resembles Protocol-2, differing primarily in whether both training and testing data from the source domains are used to train the model.
These standardized protocols offer a framework for assessing the generalizability of models across a diverse range of domains.
It is important to note that all the ablation studies are conducted under Protocol-1.

\textbf{Implementation details.}
All our experiments are conducted on an NVIDIA GeForce RTX 3090 GPU.
ViT-B/16 is employed as our backbone network, with ``B'' indicating the base ViT architecture and ``16'' specifying the patch size utilized within the model.
The training of image encoder is performed for a total of 60 epochs, with a batch size of 128.
All parameters are referenced from the CLIP-ReID~\cite{li2023clip}, while specific parameters unique to our method are further analyzed in Sec.~\ref{sec:param}.

\begin{table*}[ht]

\centering
\caption{Comparison with state-of-the-art methods under Protocol-1. The ``\({\ast}\)'' indicates results obtained based on the implementation from open-source code. The \textbf{\textcolor{red}{blod red}} indicates the best result, the \textcolor{blue}{blue} represents the second best result. }
    \setlength\tabcolsep{5pt}
\begin{tabular}{lcp{0.6cm}<{\centering} p{0.6cm}<{\centering}p{0.6cm}<{\centering} p{0.6cm}<{\centering}p{0.6cm}<{\centering} p{0.6cm}<{\centering}p{0.6cm}<{\centering} p{0.6cm}<{\centering}p{0.7cm}<{\centering} p{0.7cm}<{\centering}}
    \Xhline{1px}
    \multicolumn{1}{c}{\multirow{2}*{Method}} &\multirow{2}*{Reference} & \multicolumn{2}{c}{PRID} & \multicolumn{2}{c}{GRID} & \multicolumn{2}{c}{VIPeR} & \multicolumn{2}{c}{iLIDs} & \multicolumn{2}{c}{Average} \\
       &  &    mAP & R1 & mAP & R1 & mAP & R1 & mAP & R1 & mAP & R1\\
     \hline
        \multicolumn{12}{c}{CNN-based}\\
\cline{1-12}
     QAConv\( _{50}\)& ECCV\(_{2020}\) & 62.2 & 52.3 & 57.4 & 48.6 & 66.3 & 57.0 & 81.9 & 75.0 & 67.0 & 58.2\\
       M\( ^3\)L & CVPR\(_{2021}\) & 65.3 & 55.0 & 50.5 & 40.0 & 68.2 & 60.8 & 74.3 & 65.0 & 64.6 & 55.2\\
      MetaBIN &CVPR\(_{2021}\) & 70.8 & 61.2 & 57.9 & 50.2 & 64.3 & 55.9 & 82.7 & 74.7 & 68.9 & 60.5\\
       META &ECCV\(_{2022}\) & 71.7 & 61.9 & 60.1 & 52.4 & 68.4 & 61.5 & 83.5 & 79.2 & 70.9 & 63.8\\
           ACL & ECCV\(_{2022}\) &73.4 & 63.0 &65.7 & 55.2 & 75.1 & 66.4 & 86.5 & 81.8 & 75.2 &66.6\\
        GMN & TCSVT\(_{2024}\) & 75.4 &66.0& 64.8 &54.4 &77.7& 69.0 & - & - & - & -  \\
        ReFID & TOMM\(_{2024}\) & 71.3 & 63.2 & 59.8  &56.1 & 68.7 & 60.9 &  84.6 &81.0 & 71.1 & 65.3 \\

    \cline{1-12}
         \multicolumn{12}{c}{ViT-based}\\
    \cline{1-12}
    ViT-B\(^{\ast}\) & ICLR\(_{2021}\) &63.8	&52.0	&56.0	&44.8	&74.8	&65.8	&76.2	&65.0	&67.7	&56.9\\
       TransReID\(^*\)  & ICCV\(_{2021}\) &68.1 &	59.0 &	60.8  &	49.6 &	69.5&60.1 &	79.8 & 68.3 & 69.6 &59.3 \\
      CLIP-ReID\(^{\ast}\)  & AAAI\(_{2023}\)  & 68.3 &	57.0 &	58.2 &	48.8	 &69.3 &	60.1 &	83.4	 &75.0 &	69.8	 &60.2\\
       PAT\(^{\ast}\)  & ICCV\(_{2023}\)  &57.9	&46.0 &	54.5&	45.6&	67.8	&60.1	&78.1	&66.7	&64.6	& 54.6\\
      DSM+SHS  & MM\(_{2023}\)  &78.1 &  \textcolor{blue}{69.7} & 62.1& 53.4 & 71.2&62.8  &84.8 & 77.8&74.1  &66.0\\
      \rowcolor[rgb]{ .925,  .925,  .925}\textbf{DFGS\(_{T(\cdot)}\)} & This paper & \textcolor{blue}{78.8} & 69.0 &\textcolor{blue}{73.6} & \textcolor{blue}{66.4} & \textbf{\textcolor{red}{84.6}} & \textbf{\textcolor{red}{78.8}} & \textcolor{blue}{92.4} & \textcolor{blue}{88.3} & \textcolor{blue}{82.4} & \textcolor{blue}{75.6}\\
      \rowcolor[rgb]{ .925,  .925,  .925}\textbf{DFGS\(_{I(\cdot)}\)} & This paper & \textbf{\textcolor{red}{78.6}} & \textbf{\textcolor{red}{72.0}} &\textbf{\textcolor{red}{78.4}} & \textbf{\textcolor{red}{69.6}} & \textcolor{blue}{81.3} & \textcolor{blue}{74.4} & \textbf{\textcolor{red}{93.5}} & \textbf{\textcolor{red}{90.0}} & \textbf{\textcolor{red}{83.0}} & \textbf{\textcolor{red}{76.5}}\\
     \Xhline{1px}
\end{tabular}
\label{tab:1}
\end{table*}

\subsection{Comparison with State-of-the-art Methods}
We compare our method with state-of-the-art (SOTA) methods in generalizable person re-identification, including SNR~\cite{DBLP:conf/cvpr/JinLZ0Z20}, QAConv\( _{50}\)~\cite{Liao_Shao_2020}, M\( ^{3}\)L~\cite{Zhao_Zhong_Yang_Luo_Lin_Li_Sebe_2021}, MetaBIN~\cite{Choi_Kim_Jeong_Park_Kim_2021}, META~\cite{xu2022mimic}, ACL~\cite{Zhang_Dou_Yunlong_Li} and IL~\cite{Tan_Wang_Ding_Gong_Jia_2022}.
In addition to the CNN-based architectures mentioned above, we actively explored comparisons with models utilizing ViT as a backbone, including ViT~\cite{ViT}, TransReID~\cite{he2021transreid}, CLIP-ReID~\cite{li2023clip}, PAT~\cite{ni2023part}, DSM+SHS~\cite{Li2023StyleControllableGP}, ReFID~\cite{peng2024refid}, GMN~\cite{qi2024generalizable} and so on.
Since DukeMTMC has been withdrawn, we do not present the evaluation results for DukeMTMC. It's worth noting that methods like PAT~\cite{ni2023part} follow protocols that involve the DukeMTMC dataset, so we conducted experiments using their open-source code based on the new protocols mentioned in Sec.~\ref{sec:protocols}.
As shown in Tab.~\ref{tab:1} and Tab.~\ref{tab:2}, our method consistently outperforms others in all three protocols, as evidenced by the higher average mAP and R1 results.
This serves as validation for the efficacy of our method in enhancing the model's generalization to unseen domains.

It is worth noting that both ACL and META utilized four GPUs for training their models, whereas our method only requires one GPU.
Under this setup, our method exhibits superior performance compared to ACL.
In Protocol-1, our method outperforms ACL by \textbf{+7.8\%} (83.0\% vs. 75.2\%) on mAP.
Similarly, in Protocol-2, our method outperforms ACL by \textbf{+8.3\%} (53.6\% vs. 45.3\%) on mAP.
Furthermore, in Protocol-3, our method achieves a mAP of 55.3\%, which is \textbf{+6.0\%} higher than ACL.

Comparing our method to other SOTA methods in the context of Protocol-1, it becomes evident that our model exhibits superior performance.
\textbf{ReFID}, which is one of the top-performing methods in this comparison, achieved an mAP of 71.1\% and an R1 of 65.3\%, while \textbf{DSM+SHS}, another competitive method, achieved an mAP of 74.1\% and an R1 of 66.0\%.
In both mAP and R1, our method outperforms these models with an mAP of \textbf{83.0\%} and an R1 of \textbf{76.5\%}.

In the context of Protocol-2 and Protocol-3, our method outperforms other SOTA methods in terms of average mAP and average R1.
Especially under Protocol-2, our method exhibits results that are unmatched by other methods.
It achieved an impressive average mAP of \textbf{53.6\%} and a remarkable R1 of \textbf{67.1\%}, demonstrating its exceptional effectiveness.
In comparison, the prior state-of-the-art models, such as \textbf{GMN} with an average mAP of 46.6\% and an R1 of 60.0\%, showed notable but comparatively lower performance.
Furthermore, in experiments under Protocol-3, while not showcasing the comprehensive superiority observed in Protocol-2, our method still achieves the highest evaluations in terms of average mAP and average R1.
In Protocol-3, the results highlight our method's effectiveness in various scenarios, achieving an impressive average mAP of \textbf{55.3\%} and a remarkable R1 of \textbf{68.3\%}.
Compared to TOMM2024's SOTA method \textbf{ReFID}, which achieved 46.2\% and 58.5\% in mAP and R1 respectively, our method demonstrates a significant improvement.

Overall, the experimental results highlight the exceptional performance of our method, showing its superiority over other SOTA methods.
Whether applied to the image encoder with DFGS\(_{I(\cdot)}\) or the text encoder with DFGS\(_{T(\cdot)}\), our method consistently exhibits high mAP and R1 values across multiple datasets, further emphasizing its effectiveness.

\begin{table*}[t]
  \centering
  \caption{Comparison with state-of-the-art methods under Protocol-2 and Protocol-3. The ``\({\ast}\)'' indicates results obtained based on the implementation from open-source code, and the \uwave{wavy line} indicates that the method is based on a ViT backbone. The \textbf{\textcolor{red}{blod red}} indicates the best result, the \textcolor{blue}{blue} represents the second best result. }

  \resizebox{\linewidth}{!}{
  \begin{tabular}{c|l|l|  c c| c c| c c| c c}
        \Xhline{1px}
     \multicolumn{1}{c|}{\multirow{2}*{Setting}} &\multicolumn{1}{c|}{\multirow{2}*{Method}} &\multirow{2}*{Reference}& \multicolumn{2}{c|}{M+MS+CS\(\rightarrow\)C3} & \multicolumn{2}{c|}{M+CS+C3\(\rightarrow\)MS} & \multicolumn{2}{c|}{MS+CS+C3\(\rightarrow\)M} & \multicolumn{2}{c}{\multirow{1}*{Average}} \\
         &  &  &mAP   & R1    &mAP   & R1    &mAP   & R1    &mAP   &R1
%        \hline
%        \multicolumn{10}{c}{Protocol-2}\\
\\
\hline

        \multirow{12}*{P-2}  &  QAConv\(_{50}\) &ECCV\(_{2020}\) & 25.4  & 24.8  & 16.4  & 45.3  & 63.1  & 83.7  & 35.0    & 51.3 \\
         &   M3L  &CVPR\(_{2021}\)& 34.2  & 34.4  & 16.7  & 37.5  & 61.5  & 82.3  & 37.5  & 51.4 \\
         &   MetaBIN &CVPR\(_{2021}\) & 28.8  & 28.1  & 17.8  & 40.2  & 57.9  & 80.1  & 34.8  & 49.5 \\
         &   \uwave{ViT-B\(^*\)} & ICLR\(_{2021}\)& 36.5 & 35.8 & 20.5 & 42.7 & 59.2 & 78.3 & 38.7 & 52.3\\
       &   \uwave{TranReID\(^*\)} & ICCV\(_{2021}\) & 36.5 & 36.1 & 23.2 & 46.3 & 59.9 & 79.8 &39.9  & 54.1  \\
       &   META & ECCV\(_{2022}\) & 36.3  & 35.1  & 22.5  & 49.9  & 67.5  & 86.1  & 42.1  & 57.0 \\
       &     ACL  & ECCV\(_{2022}\) & 41.2  & 41.8  & 20.4  & 45.9  & 74.3  & 89.3  & 45.3  & 59.0 \\
%       &     DFF  & CVPR\(_{2023}\) & 41.3   & 41.1  &  25.1  & 50.5  & 71.1  &  87.1 & \textcolor{blue}{45.8} & 59.6 \\

       &    \uwave{CLIP-ReID\(^*\)}  & AAAI\(_{2023}\) & 42.1   & 41.9   & 26.6   & 53.1   & 68.8 & 84.4 & 45.8   & 59.8\\
       &   ReFID & TOMM\(_{2024}\) & 33.3 & 34.8 & 18.3  &39.8 & 67.6 & 85.3  & 39.7 & 53.3\\

        &   GMN & TCSVT\(_{2024}\) & 43.2 & 42.1 &  24.4 &  50.9 &  72.3 &  87.1 &  46.6 &  60.0 \\
       \rowcolor[rgb]{ .949,  .949,  .949}&    \uwave{\textbf{DFGS\(_{T(\cdot)}\)}}   & This paper &   \textcolor{blue}{45.5}   &   \textcolor{blue}{43.7}   & \textcolor{blue}{30.6}   &   \textcolor{blue}{59.4}  &     \textcolor{blue}{77.0}  & \textcolor{blue}{89.6}  & \textcolor{blue}{51.0} & \textcolor{blue}{64.2}\\
       \rowcolor[rgb]{ .949,  .949,  .949}&    \uwave{\textbf{DFGS\(_{I(\cdot)}\)}}   & This paper &  \textbf{\textcolor{red}{50.4}}   &  \textbf{\textcolor{red}{51.1}}   &    \textbf{\textcolor{red}{31.5}}    &   \textbf{\textcolor{red}{59.7}}  &    \textbf{\textcolor{red}{79.0}}  &   \textbf{\textcolor{red}{90.5}}  &  \textbf{\textcolor{red}{53.6}} &  \textbf{\textcolor{red}{67.1}}
       \\
     \hline
%        \multicolumn{10}{c}{Protocol-3}\\
%\hline
      \multirow{13}*{P-3}  &   QAConv\(_{50}\) & ECCV\(_{2020}\) & 32.9  & 33.3  & 17.6  & 46.6  & 66.5  & 85.0    & 39.0    & 55.0 \\
       &     M\(^3\)L &CVPR\(_{2021}\)  & 35.7  & 36.5  & 17.4  & 38.6  & 62.4  & 82.7  & 38.5  & 52.6 \\
       &     MetaBIN &CVPR\(_{2021}\) & 43.0    & 43.1  & 18.8  & 41.2  & 67.2  & 84.5  & 43.0    & 56.3 \\
      &   \uwave{ViT-B\(^*\)} & ICLR\(_{2021}\)& 39.4 &39.4 & 20.9 & 43.1 & 63.4 & 81.6 & 41.2 & 54.7\\
       &   \uwave{TranReID\(^*\)}  & ICCV\(_{2021}\) & 44.0 & 45.2 & 23.4 & 46.9 & 63.6 & 82.5 & 43.7 &  58.2 \\
      &   META &ECCV\(_{2022}\) & 47.1  & 46.2  & 24.4  & 52.1  & 76.5  & 90.5  & 49.3  & 62.9 \\
       &     ACL  &ECCV\(_{2022}\) & 49.4  & 50.1 & 21.7  & 47.3  & 76.8  & 90.6 & 49.3  & 62.7 \\
       &    META+IL  &TMM\(_{2023}\) & 48.9 & 48.8 & 26.9  &54.8   & 78.9 & 91.2 & 51.6 & 64.9 \\
%      &    DFF  & CVPR\(_{2023}\) &   \textbf{\textcolor{red}{51.1}}  &  \textbf{\textcolor{red}{51.2}}  &  25.3  & 51.8  & \textbf{\textcolor{red}{81.0}}   &  \textbf{\textcolor{red}{92.3}} & 52.6 & 65.1 \\

       &    \uwave{CLIP-ReID\(^*\)}  & AAAI\(_{2023}\) & 44.9 & 45.8 & 26.8 & 52.6 & 67.5 & 83.4 & 46.4 &60.6\\
        &   ReFID & TOMM\(_{2024}\) & 45.5 &  44.2 &  20.6 &  43.3 &  72.5 &  87.9 & 46.2 & 58.5\\
        &   GMN & TCSVT\(_{2024}\) & 49.5 &  \textcolor{blue}{50.1} &  24.8 &  51.0 &  75.9 &  89.0 &  50.1 &  63.4\\
       \rowcolor[rgb]{ .949,  .949,  .949} &    \uwave{\textbf{DFGS\(_{T(\cdot)}\)}}  & This paper& \textcolor{blue}{50.0} &  49.6  &    \textcolor{blue}{32.0}   &  \textcolor{blue}{60.9}  &  \textcolor{blue}{79.3}  &  \textcolor{blue}{91.3}  &  \textcolor{blue}{53.8}   &  \textcolor{blue}{67.3} \\
       \rowcolor[rgb]{ .949,  .949,  .949} &    \uwave{\textbf{DFGS\(_{I(\cdot)}\)}}  & This paper& \textbf{\textcolor{red}{51.6}} &  \textbf{\textcolor{red}{51.3}}  &   \textbf{\textcolor{red}{33.4}}   &  \textbf{\textcolor{red}{62.0}}  & \textbf{\textcolor{red}{81.0}}  & \textbf{\textcolor{red}{91.6}}  & \textbf{\textcolor{red}{55.3}}   & \textbf{\textcolor{red}{68.3}} \\
       \Xhline{1px}
    \end{tabular}
    }
\label{tab:2}

\end{table*}%

\subsection{Ablation Studies}
\label{sec:abs}
To thoroughly explore the impact of our method, we conduct a series of ablation experiments, specifically designed to assess their effects comprehensively.
The detailed outcomes of these experiments are presented in Tab.~\ref{tab:3}.

\textbf{The effectiveness of the shuffle operation before sampling.}
The experimental data in the table demonstrates a significant improvement when applying our method compared to the baseline.
Specifically, ``s.'' indicates that before sampling, a shuffle operation is performed on the graph nodes, as detailed in Alg.~\ref{alg:1}.
The presence or absence of the shuffle operation results in noticeable differences.
Since we maintain a pair-wise distance matrix, which encapsulates the similarity between samples, this matrix remains unchanged during the subsequent training of the image encoder.
If we shuffle the nodes based on the graph structure generated from this matrix, it would lead to a significant similarity among iterations of training the image encoder (the same hard sample pairs would frequently be grouped into the same mini-batch).
This would diminish the diversity of metric learning.
However, incorporating the shuffle operation not only does not increase training time, but also enhances the diversity of sample pairs, thereby improving the model's discriminative capability.

\begin{wraptable}{R}{0.6\columnwidth}
\caption{Ablation studies of our method. ``s.''  represents the shuffle operation for distance matrix before each epoch. All experiments are conducted under the protocol-1 setting.}
        \centering
      \begin{tabular}{l| c | c |cc | c}
    \Xhline{1px}

     \multicolumn{1}{c|}{\multirow{2}*{Method}} & \multirow{2}*{s.}  & \multirow{2}*{\(\mathcal{L}_{id}\)}  &  \multicolumn{2}{c|}{Average}  & Training time  \\
    &  &   &   mAP & R1 & (min)\\
    \hline
          \multirow{2}*{Baseline} & \multirow{2}*{-}  & $ \checkmark $ &	69.8	 &60.2 & 490\\
      &   & $ \times $&	68.2 & 59.5   & 482\\
     \hline
      \multirow{2}*{+ DFGS\(_{T(\cdot)}\)} & \multirow{2}*{$ \times $}  &  $ \checkmark $&78.6 &69.2 & 470\\
                                &  & $ \times $  &  79.8 &71.3 & 463\\
    \hline
      \multirow{2}*{+ DFGS\(_{T(\cdot)}\)}& \multirow{2}*{$ \checkmark $} & $ \checkmark $&80.8 &71.2 & 466\\
       &  & $ \times $ & 82.4 & 75.6 & 459\\
    \hline
        \multirow{2}*{+ DFGS\(_{I(\cdot)}\)} & \multirow{2}*{$ \times $}  & $ \checkmark $ & 81.3 & 74.8 & 537\\
       &  & $ \times $ &82.8 &75.6 & 529\\
    \hline
        \multirow{2}*{+ DFGS\(_{I(\cdot)}\) ~}& \multirow{2}*{$ \checkmark $} & $ \checkmark $ & 82.2 & 74.9 & 539\\
       &  & $ \times $ & 83.0 &76.5 & 530\\
    \Xhline{1px}
    \end{tabular}
        \label{tab:3}
\end{wraptable}

\textbf{The effectiveness of DFGS\(_{T(\cdot)}\) and DFGS\(_{I(\cdot)}\).}
The distance metric matrix of DFGS\(_{I(\cdot)}\) is derived from the image encoder.
On one hand, this approach allows for obtaining challenging samples for the current model before each training round, thereby incrementally enhancing the model's discriminative capability.
However, this method can increase the overall training time due to the time overhead incurred by acquiring the distance matrix before each training round.
Interestingly, when the distance metrics are computed from features obtained from the text encoder, we observed the lowest overall training time while still achieving highly competitive results.
This finding highlights a crucial balance between efficiency and performance.
Not only does it validate the \textbf{effectiveness} of DFGS in improving model accuracy, but it also underscores the \textbf{efficiency} of applying DFGS to text features, which can significantly reduce computational overhead without compromising the quality of the results.
Overall, the application of DFGS to both image and text encoders exemplifies a powerful strategy for enhancing model performance.

\textbf{Remark 3:}
At present, most person ReID methods combine triplet loss and ID loss for extracted feature.
However, our designed DFGS method is specifically tailored for triplet loss, meaning that during sampling, we aim to fill a minibatch with hard-to-discriminate triplets.
This approach maximizes the potential of hard samples to enhance the model's discriminative capability.
When triplet loss and ID loss are used simultaneously, it may limit the full potential of hard samples.

\subsection{Further Analysis}

\textbf{Different samplers.}
We conduct experiments using three samplers, namely PK, GS~\cite{Liao_Shao_2021} and DFGS, under the same configuration following the Protocol-1 settings.
It is worth noting that, for the sake of fair comparison, we strive to maintain a similar number of iterations for each epoch of the three samplers.
Furthermore, the update frequency for both GS and DFGS has been kept uniform, thereby guaranteeing the fairness of the experiment.
We conduct experiments using ResNet, ViT and CLIP as backbone to study the effects of three different samplers separately.
The experimental results are shown in Tab.~\ref{tab:4}.
It can be observed that our method exhibits superiority compared to PK and GS.
In experiments based on ResNet as the backbone network, GS achieves a mAP 1.4\% higher than PK and R1 0.8\% higher.
Impressively, our DFGS surpasses GS in mAP by 3.3\% and R1 by 3.3\%.
In experiments based on ViT as the backbone, the mAP results using GS sampling are 3.9\% higher than PK, and our DFGS outperforms GS by an additional 1.8\%.
In experiments based on CLIP, the mAP results using GS sampling are 8.9\% higher than PK, and our DFGS outperforms GS by an additional 4.3\%.
It should be noted that since the PK sampler does not involve the graph construction process, its runtime is lower than that of GS and DFGS while maintaining a similar number of iterations.
However, it is also evident that DFGS, when running on the text encoder, achieves competitive runtime while ensuring a certain level of performance, which is one of the significant advantages of our method.
In conclusion, our sampling method outperforms GS and is particularly more effective than PK in person ReID.

\begin{table*}[ht]
\centering
\caption{The results of three different samplers are obtained on ViT and CNN, respectively. The ``\(\star\)"  represents the result based on open-source code implementation. The \textbf{\textcolor{red}{bold red}} indicates the best result and \textcolor{blue}{blue} represents the second best result.}
  \resizebox{\linewidth}{!}{
\begin{tabular}{l|l|cc| cc| cc| cc| cc| c}
    \Xhline{1px}
\multicolumn{1}{c|}{\multirow{3}*{Backbone}} & \multicolumn{1}{c|}{\multirow{3}*{Method}}&\multicolumn{10}{c|}{Source domain: M+C2+C3+CS} & Training\\
    \cline{3-12}
 & &  \multicolumn{2}{c|}{Target: PRID} & \multicolumn{2}{c|}{Target: GRID} & \multicolumn{2}{c|}{Target: VIPeR} & \multicolumn{2}{c|}{Target: iLIDs} & \multicolumn{2}{c|}{Average} & Time \\

%    \cline{4-13}
       &  &    mAP & R1 & mAP & R1 & mAP & R1 & mAP & R1 & mAP & R1 & (min)\\
     \hline
      \multirow{3}*{ResNet\(_{50}\)~}
          & PK  & 45.5 & 33.0 & 54.4 & 44.8 & \textcolor{blue}{64.5} & \textcolor{blue}{53.5} & \textcolor{blue}{78.5} & \textbf{\textcolor{red}{70.0}} & 60.7 & 50.3 & 205\\
          & GS\(^\star\)  & \textcolor{blue}{51.3} & \textcolor{blue}{38.0} & \textcolor{blue}{56.8} & \textcolor{blue}{45.6} & 62.9 & 52.5 & 77.3 & \textcolor{blue}{68.3} & \textcolor{blue}{62.1} & \textcolor{blue}{51.1} &263\\
         & DFGS\(_{I(\cdot)}\) & \textbf{\textcolor{red}{56.9}} & \textbf{\textcolor{red}{46.0}} & \textbf{\textcolor{red}{58.9}} & \textbf{\textcolor{red}{48.8}} & \textbf{\textcolor{red}{67.2}} & \textbf{\textcolor{red}{56.0}} & \textbf{\textcolor{red}{78.6}} & 66.7 & \textbf{\textcolor{red}{65.4}} & \textbf{\textcolor{red}{54.4}} & 267\\
     \hline
      \multirow{3}*{ViT-B} & PK  & 63.8 & 52.0 & 56.0 & 44.8 & 74.8 & 65.8 & 76.2 & 65.0 & 67.7 & 56.9 & 312\\
     & GS\(^\star\)  & \textcolor{blue}{68.8} & \textcolor{blue}{58.0} & \textcolor{blue}{58.7} & \textcolor{blue}{50.4} & \textcolor{blue}{76.8} & \textcolor{blue}{68.4} & \textbf{\textcolor{red}{81.9}} & \textbf{\textcolor{red}{73.3}} & \textcolor{blue}{71.6} & \textcolor{blue}{62.5}& 421\\
           & DFGS\(_{I(\cdot)}\)  & \textbf{\textcolor{red}{73.4}} & \textbf{\textcolor{red}{63.0}} & \textbf{\textcolor{red}{61.1}} & \textbf{\textcolor{red}{52.8}} & \textbf{\textcolor{red}{78.2}} & \textbf{\textcolor{red}{70.9}} & \textcolor{blue}{80.8} & \textcolor{blue}{71.7} & \textbf{\textcolor{red}{73.4}} & \textbf{\textcolor{red}{64.6}}& 405 \\
     \hline
      \multirow{4}*{CLIP} & PK  & 68.3 &	57.0 &	58.2 &	48.8	 &69.3 &	60.1 &	83.4	 &75.0 &	69.8	 &60.2 & 490\\
          & GS\(^\star\)  & \textbf{\textcolor{red}{79.0}} & \textcolor{blue}{71.0} & 69.5 & 58.4 & 79.2 & 71.2 & 87.2 & 80.0 & 78.7 & 70.2 & 525\\
            &DFGS\(_{T(\cdot)}\)  & \textcolor{blue}{78.8} & 69.0 &\textcolor{blue}{73.6} & \textcolor{blue}{66.4} & \textbf{\textcolor{red}{84.6}} & \textbf{\textcolor{red}{78.8}} & \textcolor{blue}{92.4} & \textcolor{blue}{88.3} & \textcolor{blue}{82.4} & \textcolor{blue}{75.6} & 459\\
            & DFGS\(_{I(\cdot)}\)  & 78.6 & \textbf{\textcolor{red}{72.0}} &\textbf{\textcolor{red}{78.4}} & \textbf{\textcolor{red}{69.6}} & \textcolor{blue}{81.3} & \textcolor{blue}{74.4} & \textbf{\textcolor{red}{93.5}} & \textbf{\textcolor{red}{90.0}} & \textbf{\textcolor{red}{83.0}} & \textbf{\textcolor{red}{76.5}} & 530\\
    \Xhline{1px}
\end{tabular}}
\label{tab:4}
\end{table*}

\begin{figure}[ht]
    \begin{subfigure}{0.32\linewidth}
    \centering
    \includegraphics[width=0.99\linewidth]{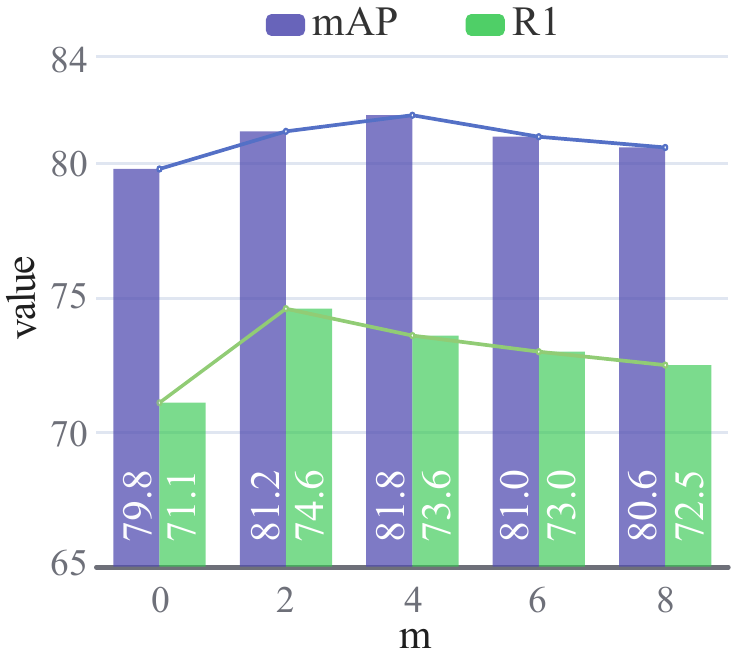}
    \caption{}
    \label{fig:4a}
    \end{subfigure}
\hfill
  \begin{subfigure}{0.32\linewidth}
    \centering
    \includegraphics[width=0.99\linewidth]{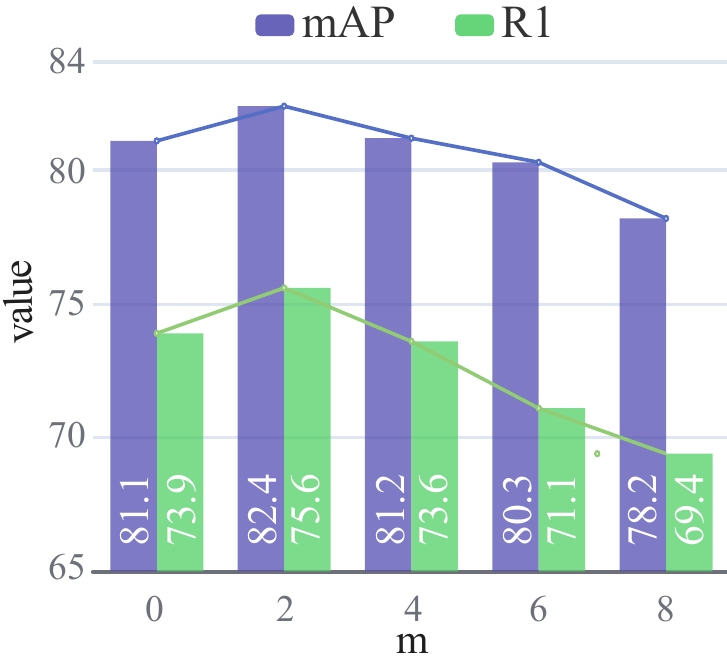}
    \caption{}
    \label{fig:4b}
\end{subfigure}
\hfill
  \begin{subfigure}{0.32\linewidth}
    \centering
    \includegraphics[width=0.99\linewidth]{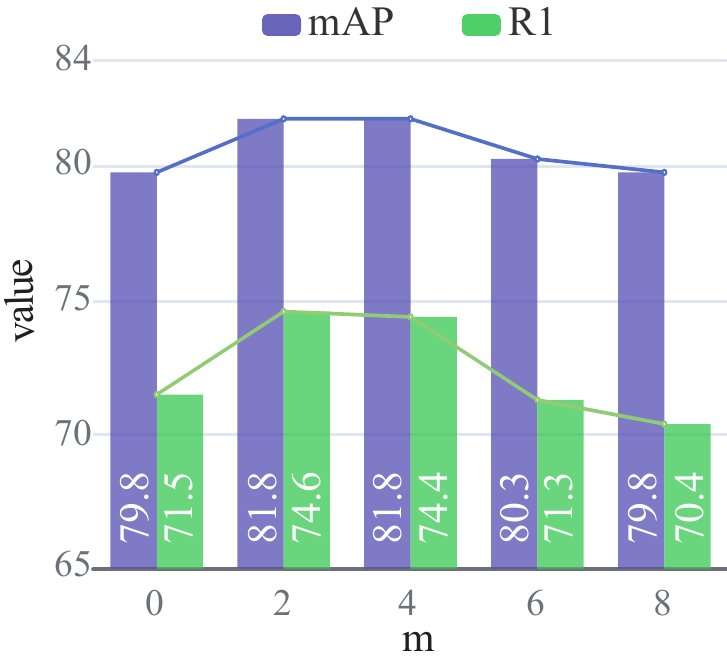}
    \caption{}
    \label{fig:4c}
\end{subfigure}
    \caption{Parameter analysis: (a)(b)(c) represent the cases where \( k \) is 5, 10 and 15.}
    \label{fig:4}
\end{figure}

\textbf{Parameter analysis.}
\label{sec:param}
Experiments are conducted regarding the hyperparameter \( m\) and \( k\) in Eq.~\eqref{gpv2}.
As shown in the Fig.~\ref{fig:4}, we studied three sets of different parameters for \( k \), which are 5, 10, and 15.
For each set, we further divided different initial values for \( m \), which are 0, 2, 4, 6, and 8.

The horizontal axis of the three figures represents different parameters \( m \), while Fig.~\ref{fig:4a}, Fig.~\ref{fig:4b} and Fig.~\ref{fig:4c} respectively correspond to the cases where \( k \) is 5, 10, and 15.
It can be observed that these three figures generally exhibit a pattern: they initially increase from 0, reach their peak around \( m = 2 \), and then decrease as \( m \) continues to increase.
We analyzed that in the training data, there exist highly similar or nearly identical samples.
Mining these types of samples may pose significant challenges for the model in terms of discrimination, leading to a decrease in model performance.
Avoiding these most difficult samples can actually improve the model's discriminative ability.
Furthermore, through the comparative analysis of the three figures, we found that the highest results are achieved when \( k = 10 \).
We believe that when \( k \) is small, the samples are all difficult samples and lack generalization; when \( k \) is extremely large, the proportion of difficult samples in the overall dataset may be low.
Both of these situations may lead to a decrease in model performance.
By controlling \( k \) to be around 10, we can fully utilize difficult samples to improve discriminative ability, while also maintaining model performance with a sufficient number of general samples.

\textbf{Experiments on BLIP.}
To further validate the effectiveness, we attempt to apply our proposed method on BLIP.
Applying our method to BLIP is straightforward, we simply need to compute the distance metric matrix required during the sampling process using the features obtained from BLIP's text encoder.
As evident from the Tab.~\ref{tab:5}, applying our method to BLIP could effectively improves performance.
Our method achieves an average mAP increase of 5.2\% and an R1 increase of 6.8\%.
This indicates the potential of our method to be applied to VLMs other than CLIP and enhance model discriminative ability.

\begin{table}[ht]

  \centering
\caption{Experiments of applying our method to the BLIP model.}
  \resizebox{\linewidth}{!}{
    \begin{tabular}{c| l| cc| cc| cc| cc| cc}
    \Xhline{1px}
          \multicolumn{1}{c|}{\multirow{3}*{Source domain}} &\multicolumn{1}{c|}{\multirow{3}*{Method}} &\multicolumn{10}{c}{Target domain}\\
    \cline{3-12}
    & &   \multicolumn{2}{c|}{PRID} & \multicolumn{2}{c|}{GRID} & \multicolumn{2}{c|}{VIPeR} & \multicolumn{2}{c|}{iLIDs} & \multicolumn{2}{c}{Average} \\
    &  &      mAP & R1 & mAP & R1 & mAP & R1 & mAP & R1 & mAP & R1\\ \hline
     \multirow{3}*{M+C2+C3+CS} &   BLIP~  &     64.4   &   51.0    &    66.4   &   56.0    &    76.4   &    67.7   &    83.9   &  76.7     &   72.8    & 62.8  \\
      & + DFGS\(_{T(\cdot)}\) &  73.4 & 63.0 & 74.6 & 67.2 & 76.6 & 68.7 & 84.2 & 76.7 & 77.2\(^{\textcolor{red}{\uparrow{4.4}}}\) & 68.9\(^{\textcolor{red}{\uparrow{6.1}}}\) \\
       & + DFGS\(_{I(\cdot)}\) &  73.9 & 63.0 & 73.6 & 66.4 & 78.2 & 69.9 & 86.3 & 79.2 & 78.0\(^{\textcolor{red}{\uparrow{5.2}}}\) & 69.6\(^{\textcolor{red}{\uparrow{6.8}}}\) \\

    \Xhline{1px}
    \end{tabular}}
  \label{tab:5}%
\end{table}%

\textbf{Performance on single-source-domain person ReID task}
The datasets involved in the following experiments include: Market1501, MSMT17, CUHK02, CUHK03, and CUHK-SYSU.
We train model on one of the datasets: Market1501, MSMT17, CUHK02, CUHK03 and test on the remaining datasets separately.
From Tab.~\ref{tab:6}, it can be observed that our method achieves significant improvements in single-source domain person re-identification compared to the baseline.
Notably, when training on MSMT17, our method increased the mAP from 53.0\% to 55.2\% and the R1 from 80.0\% to 80.9\% on the Market1501 test set.
For instance, on the CUHK02 test set, training on CUHK03 resulted in an mAP increase from 55.2\% to 55.4\% and an R1 increase from 77.0\% to 77.7\%.
When training on CUHK-SYSU and testing on CUHK03, the mAP rose from 71.6\% to 73.5\% and R1 from 87.1\% to 88.3\%.
This comprehensive improvement highlights the robustness and effectiveness of our method in various single-source domain scenarios.
%These enhancements validate our methodological choices, proving our method's superiority and robustness, making it a versatile solution for various real-world applications.

%Leveraging_Vision-Language_Model_for_Hard_Sample_Mining_in_Generalizable_Person_Re-Identification_v1

\begin{table*}[ht]
\centering
\caption{The results of our method in single-source domain person re-identification.The \textcolor{blue}{blue \(_{\downarrow}\)} indicate a decrease in performance compared to the baseline, while the \textcolor{red}{red \(^{\uparrow}\)} indicate an improvement.}
  \resizebox{\linewidth}{!}{
\begin{tabular}{l | c | cc| cc| cc| cc}
    \Xhline{1px}
    \multicolumn{1}{c|}{\multirow{3}*{Method}}  & \multicolumn{1}{c|}{\multirow{3}*{Train Set}} &  \multicolumn{8}{c}{Test Set} \\
    \cline{3-10}
    &        &     \multicolumn{2}{c|}{Market1501} & \multicolumn{2}{c|}{MSMT17} & \multicolumn{2}{c|}{CUHK02} & \multicolumn{2}{c}{CUHK03} \\
%    \cline{2-3}
    &           &     mAP & R1 & mAP & R1 & mAP & R1 & mAP & R1 \\

     \hline
    Baseline& \multirow{2}*{Market1501\(_\text{all}\)}  & - & - & 21.8 & 49.9 &74.2 & 72.2 & 36.8 & 37.9\\
                     Ours &     &- & - &~21.5\(_{\textcolor{blue}{\downarrow}}\) &~49.7\(_{\textcolor{blue}{\downarrow}}\) & ~75.2\(^{\textcolor{red}{\uparrow}}\) & ~76.2\(^{\textcolor{red}{\uparrow}}\) & ~37.5\(^{\textcolor{red}{\uparrow}}\) & ~38.8\(^{\textcolor{red}{\uparrow}}\)\\

     \hline
     Baseline & \multirow{2}*{MSMT17\(_\text{all}\)} &  53.0 &80.0 & - & - &70.7 & 70.1 &35.2 &35.9 \\
                    Ours &     &~55.2\(^{\textcolor{red}{\uparrow}}\) & ~80.9\(^{\textcolor{red}{\uparrow}}\) & -& - & ~71.8\(^{\textcolor{red}{\uparrow}}\) & ~72.2\(^{\textcolor{red}{\uparrow}}\) & ~35.7\(^{\textcolor{red}{\uparrow}}\) & ~37.1\(^{\textcolor{red}{\uparrow}}\)\\

     \hline
     Baseline & \multirow{2}*{CUHK02\(_\text{all}\)}  & 44.1 & 75.6 & 17.6 & 48.9 & - & - & 60.8 & 62.9\\
                      Ours &      &~46.5\(^{\textcolor{red}{\uparrow}}\) & ~76.1\(^{\textcolor{red}{\uparrow}}\) & ~18.6\(^{\textcolor{red}{\uparrow}}\)& ~50.9\(^{\textcolor{red}{\uparrow}}\) & - & - & ~63.2\(^{\textcolor{red}{\uparrow}}\) & ~65.9\(^{\textcolor{red}{\uparrow}}\)\\

     \hline
      Baseline &\multirow{2}*{CUHK03\(_\text{all}\)}  & 55.2 & 77.0 & 18.7 & 47.8 & 95.2 & 94.6 & - &- \\
                    Ours &    & ~55.4\(^{\textcolor{red}{\uparrow}}\) & ~77.7\(^{\textcolor{red}{\uparrow}}\) & ~19.0\(^{\textcolor{red}{\uparrow}}\) & ~48.5\(^{\textcolor{red}{\uparrow}}\) & ~95.5\(^{\textcolor{red}{\uparrow}}\) & ~94.9\(^{\textcolor{red}{\uparrow}}\) & -& -\\

     \hline
      Baseline &\multirow{2}*{CUHK-SYSU\(_\text{all}\)}  & 71.6 & 87.1 & 25.1 & 52.3 & 75.2 & 72.4 & 30.5 & 29.2\\
                      Ours &    & ~73.5\(^{\textcolor{red}{\uparrow}}\) & ~88.3\(^{\textcolor{red}{\uparrow}}\) &  ~27.6\(^{\textcolor{red}{\uparrow}}\)& ~56.0\(^{\textcolor{red}{\uparrow}}\) & ~78.4\(^{\textcolor{red}{\uparrow}}\) & ~77.8\(^{\textcolor{red}{\uparrow}}\) & ~34.6\(^{\textcolor{red}{\uparrow}}\) &~34.2\(^{\textcolor{red}{\uparrow}}\)\\

    \Xhline{1px}
\end{tabular}}
\label{tab:6}
\end{table*}

\textbf{Performance on non-cross-domain person ReID task}
From Tab.~\ref{tab:7}, it is evident that our method demonstrates significant improvements over the baseline in non-cross-domain person re-identification tasks.
The average mAP increased from 78.8\% to 79.6\%, and the average R1 improved from 85.9\% to 87.2\%.
These performance gains are due to our advanced feature extraction techniques, which capture more discriminative representations, and our sophisticated sampling strategy that ensures the model trains on challenging and informative samples.
Moreover, these consistent improvements across different datasets underscore the robustness and generalizability of our method.
The combination of enhanced feature extraction and targeted sampling strategies plays a crucial role in achieving these superior results.

\begin{table}[ht]
\centering
\caption{The results of our method in non-cross-domain person re-identification. The \textcolor{red}{red \(^{\uparrow}\)} indicate an improvement.}
%  \resizebox{\linewidth}{!}{
\begin{tabular}{l | c cc cc cc c|cc}
    \Xhline{1px}
      \multicolumn{1}{c|}{\multirow{2}*{Method}}  & \multicolumn{2}{c}{M} & \multicolumn{2}{c}{MS} & \multicolumn{2}{c}{C2} & \multicolumn{2}{c|}{C3}& \multicolumn{2}{c}{Average} \\
     &     mAP & R1 & mAP & R1 & mAP & R1 & mAP & R1 & mAP & R1 \\

     \hline
     Baseline~&  85.2 &93.2 &65.5& 85.0&89.1&88.1 &75.2&77.2 & 78.8 & 85.9 \\
     \rowcolor[rgb]{0.950, 0.950, 0.950}Ours&  85.5 &93.7 &66.3& 86.4&90.0&88.9&76.6&79.6 &79.6\(^{\textcolor{red}{\uparrow{0.8}}}\) & 87.2\(^{\textcolor{red}{\uparrow{1.3}}}\) \\

    \Xhline{1px}
\end{tabular}
\label{tab:7}

\end{table}

%\textbf{Performance on other person ReID tasks}
%Our main experiments follow the three common experimental protocols used in current generalizable person re-identification tasks.
%Additionally, to further demonstrate the effectiveness of our method, we conducted experiments in both single-source domain and non-cross-domain settings.
%The datasets involved in the following experiments include: Market1501 (M), MSMT17 (MS), CUHK02 (C2), CUHK03 (C3), and CUHK-SYSU (CS).

\begin{figure}[ht]
    \begin{subfigure}{0.48\linewidth}
    \centering
    \includegraphics[width=\linewidth]{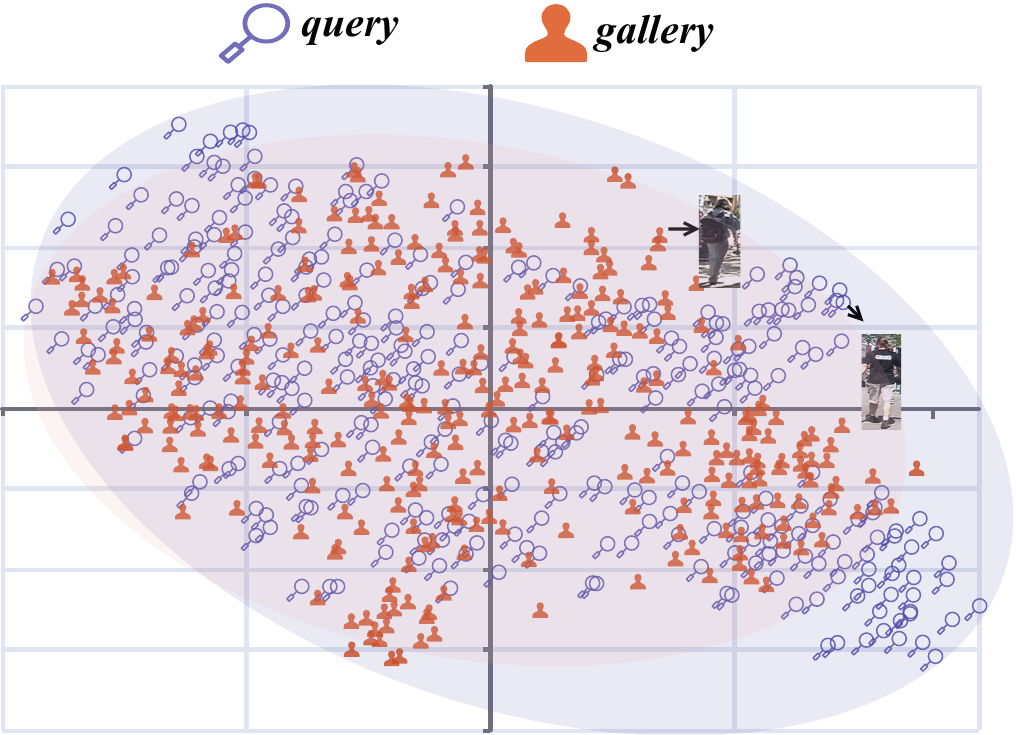}
    \caption{Baseline}
    \label{fig:5a}
    \end{subfigure}
\hfill
  \begin{subfigure}{0.48\linewidth}
    \centering
    \includegraphics[width=\linewidth]{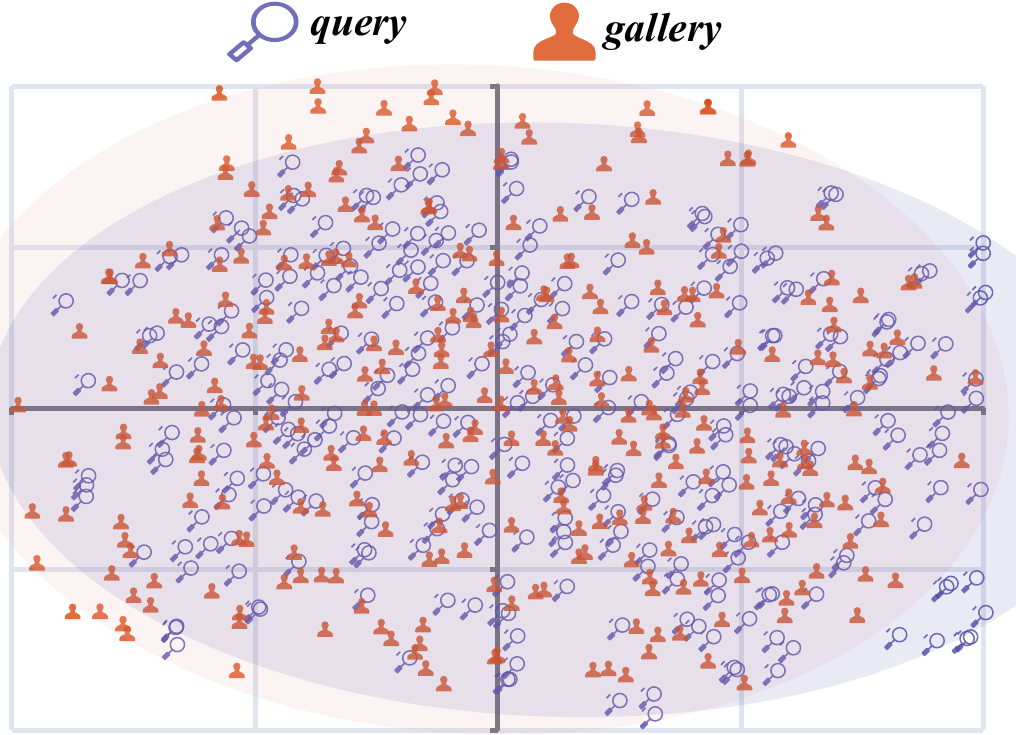}
    \caption{Ours}
    \label{fig:5b}
\end{subfigure}
    \caption{t-SNE visualization of embeddings on the VIPeR target domain dataset: (a) and (b) presents the t-SNE results for both baseline and our method, respectively. }
    \label{fig:5}
\end{figure}

\textbf{Visualization.}
To comprehensively illustrate effectiveness of our method, we visualize t-distributed Stochastic Neighbor Embedding (t-SNE)~\cite{Maaten_Hinton_2008} on the unseen target domain dataset under Protocol-1.
As shown in Fig.~\ref{fig:5}, by comparing Fig.~\ref{fig:5a} and Fig.~\ref{fig:5b}, it is evident that our method yields embedding scatter points with a smaller domain gap compared to the baseline.
This indicates that our method effectively mitigates the issue of domain generalization.

\begin{figure}[ht]
    \begin{subfigure}{0.23\linewidth}
    \centering
    \includegraphics[width=0.85\linewidth]{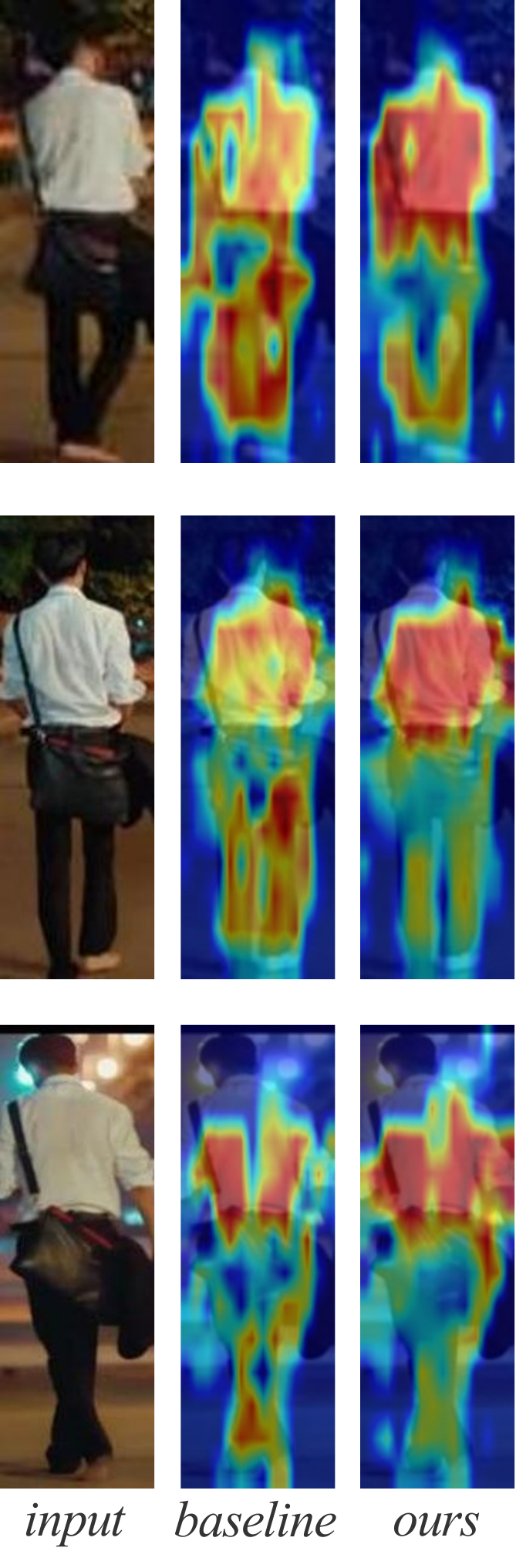}
    \caption{}
    \label{fig:6a}
    \end{subfigure}
\hfill
  \begin{subfigure}{0.23\linewidth}
    \centering
    \includegraphics[width=0.85\linewidth]{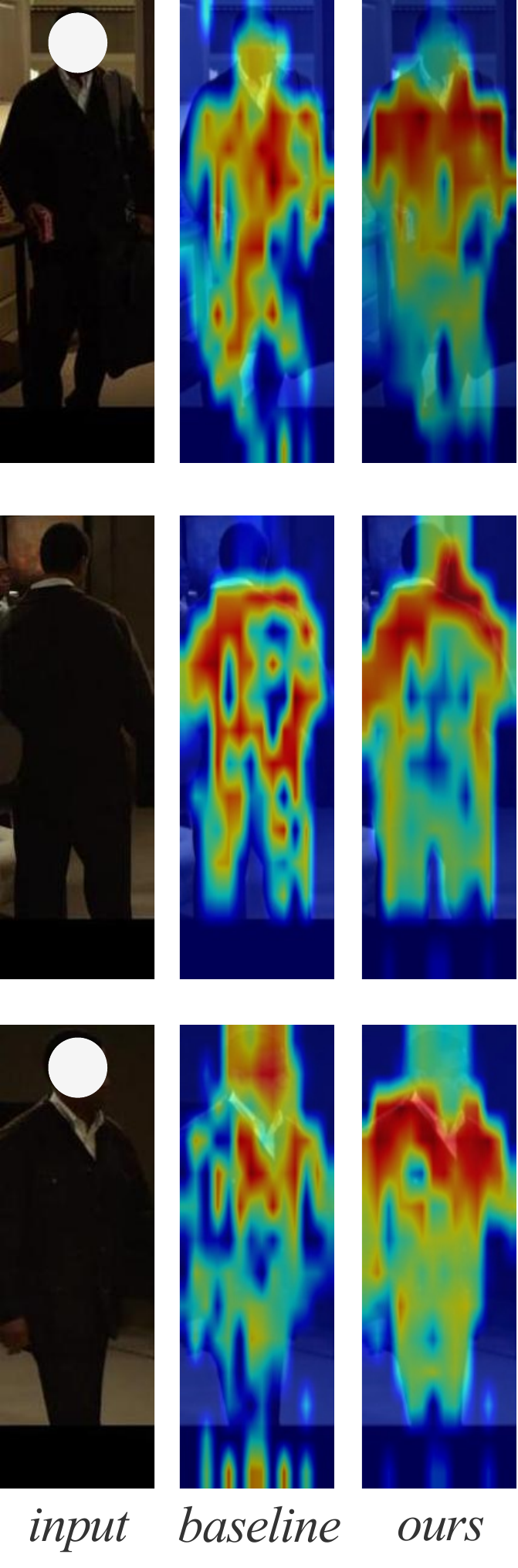}
    \caption{}
    \label{fig:6b}
\end{subfigure}
\hfill
  \begin{subfigure}{0.23\linewidth}
    \centering
    \includegraphics[width=0.85\linewidth]{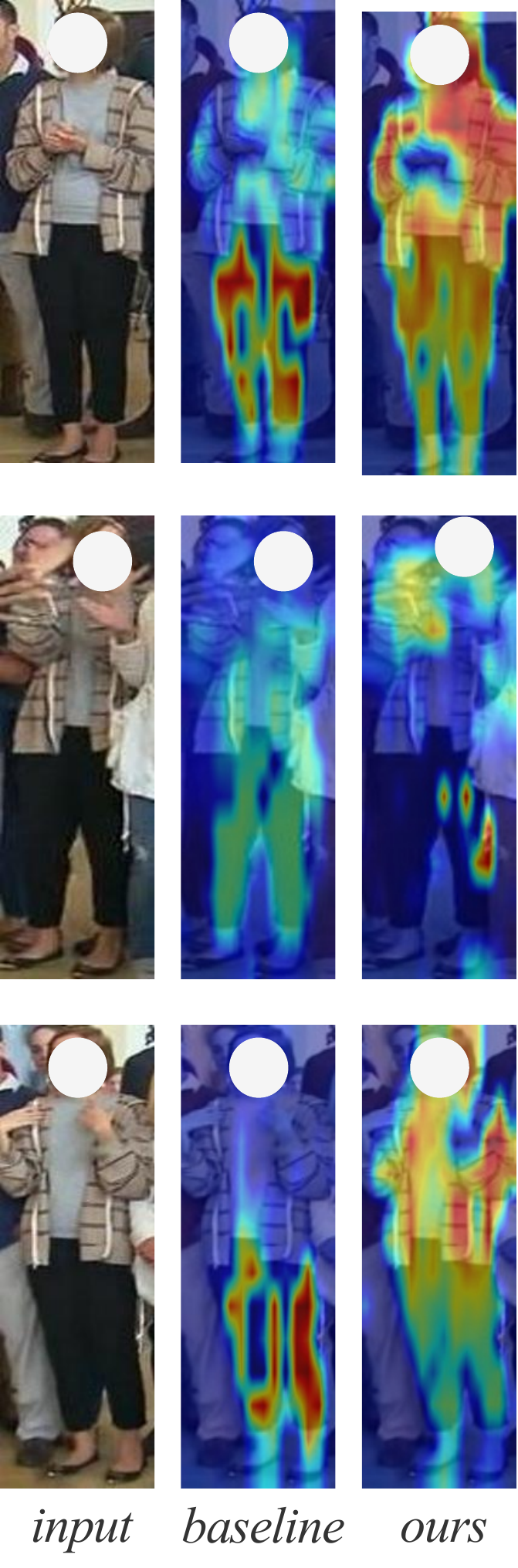}
    \caption{}
    \label{fig:6c}
\end{subfigure}
\hfill
  \begin{subfigure}{0.23\linewidth}
    \centering
    \includegraphics[width=0.85\linewidth]{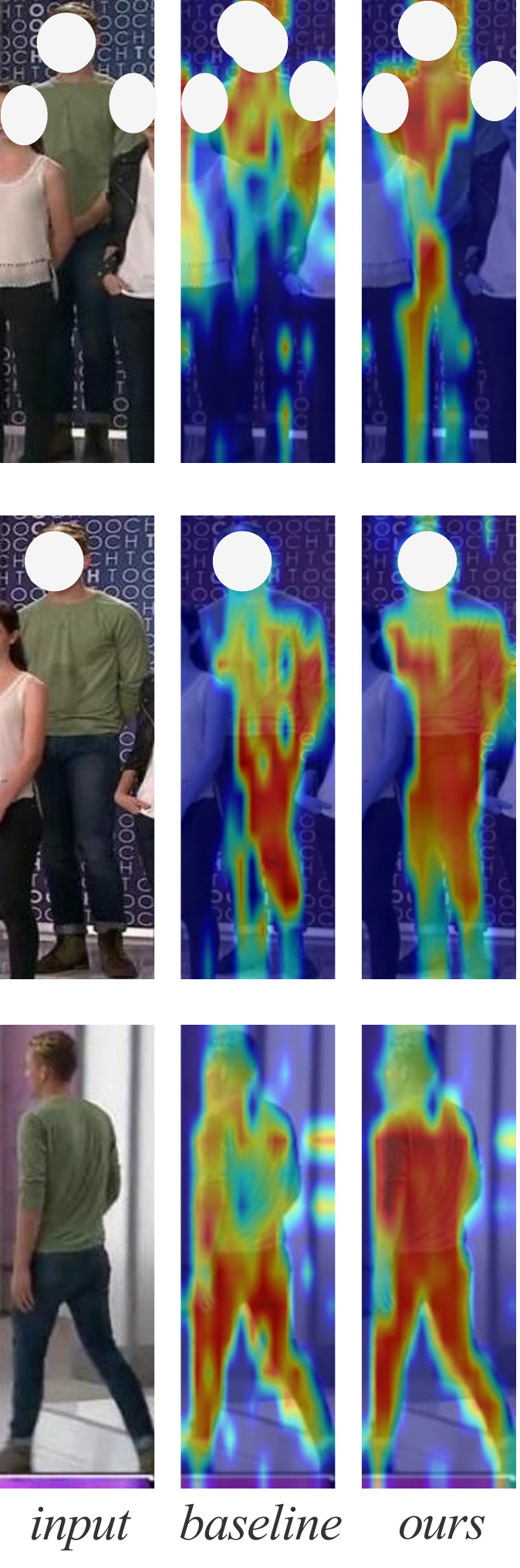}
    \caption{}
    \label{fig:6d}
\end{subfigure}

    \caption{Visualization of activation maps: (a), (b), (c) and (d) represent different individuals. The first column shows the original image, the second column displays the activation map from the baseline, and the third column presents the activation map from our method. These samples aim to encompass a wide variety of scenarios, including both daytime and nighttime conditions, as well as frontal and rear views of individuals.}
    \label{fig:6}

\end{figure}

Additionally, we utilize GradCAM~\cite{selvaraju2017grad} to generate model attention activation maps.
As shown in Fig.~\ref{fig:6}, it can clearly be seen that our method has a significantly more accurate attention scope compared to the baseline.
Moreover, as depicted in Fig.~\ref{fig:6c}, the baseline frequently tends to focus on the ``black pants'' feature, which is not a particularly distinctive discriminative feature.
Conversely, with our method applied, the model more accurately focuses on features such as the ``patterned hoodie'', which are considerably more discriminative.
This effectively demonstrates the superiority of our method over the baseline in generalizable person re-identification tasks.

\section{Conclusion}
\label{sec:conclusion}

In conclusion, we address the limitations of pre-trained vision-language models such as CLIP in generalizable person re-identification tasks by introducing DFGS (Depth-First Graph Sampler), a novel hard sample mining method.
DFGS is adaptable to both the image encoder (DFGS\(_{I(\cdot)}\)) and the text encoder (DFGS\(_{T(\cdot)}\)) within CLIP, harnessing its cross-modal learning capabilities.
Leveraging depth-first search, our method provides sufficiently challenging samples that enhance the model's ability to distinguish difficult instances.
By forming mini-batches with high discriminative difficulty, DFGS supplies the image model with more effective and challenging samples, thereby improving its capacity to differentiate between individuals.
Experimental results consistently demonstrate the superiority of DFGS over existing methods, confirming its effectiveness in enhancing CLIP's performance in generalizable person re-identification tasks.

\textbf{Limitation.}
While experiments demonstrate the superiority of our method, there are still some limitations.
DFGS is designed for CLIP and consists of DFGS\(_{I(\cdot)}\) and DFGS\(_{T(\cdot)}\), targeting the image encoder and text encoder, respectively.
As shown in the experiments presented in the paper, although DFGS\(_{T(\cdot)}\) has achieved very impressive experimental results while maintaining efficiency in runtime, there is still room for improvement compared to DFGS\(_{I(\cdot)}\)).
We believe there is potential for further optimization in refining the sampling method to effectively identify valuable hard samples while maintaining efficiency.

\bibliographystyle{ACM-Reference-Format}
\bibliography{bibliography}

\end{document}